\documentclass{article}

\usepackage{arxiv}

\usepackage[utf8]{inputenc} 
\usepackage[T1]{fontenc}    
\usepackage{hyperref}       
\usepackage{url}            
\usepackage{booktabs}       
\usepackage{amsfonts}       
\usepackage{nicefrac}       
\usepackage{microtype}      
\usepackage{lipsum}
\usepackage{graphicx}

\usepackage{amsmath, amsfonts, dsfont}
\usepackage[bbgreekl]{mathbbol}
\usepackage{amsfonts}
\usepackage{enumitem}
\usepackage{footnote}
\usepackage{float}
\usepackage{hyperref}
\usepackage{pgfplots}
\pgfplotsset{compat=1.7}
\usepackage{times}
\usepackage{multicol}
\usepackage{multirow}
\usepackage{wrapfig}
\usepackage{tikz}
\usepackage{lipsum} 
\usepackage{enumitem}
\setlist{nosep} 
\usepackage{color}
\usepackage{relsize}
\usetikzlibrary{calc}
\newcommand\floor[1]{\lfloor#1\rfloor}

\usepackage{array}
\usepackage{booktabs}
\usepackage{media9}
\usepackage{hyperref}

\usepackage{tikz}
\usepackage{forest}
\forestset{
  L1/.style={draw=black,},
  L2/.style={,edge={,line width=0.8pt}},
}

\usepackage{verbatim}
\usetikzlibrary{arrows,shapes}
\tikzstyle{format} = [draw, thin, fill=blue!20]
\tikzstyle{medium} = [ellipse, draw, thin, fill=green!20, minimum height=2.5em]
\usetikzlibrary{positioning,chains}

\usepackage[ruled,vlined]{algorithm2e}

\usepackage[utf8]{inputenc} 
\usepackage[T1]{fontenc}    
\usepackage{hyperref}       
\usepackage{url}            
\usepackage{booktabs}       
\usepackage{amsfonts}       
\usepackage{nicefrac}       
\usepackage{microtype}      
\usepackage{xcolor}         

\title{An Interaction-based Convolutional Neural Network (ICNN) Towards Better Understanding of COVID-19 X-ray Images}

\author{
 Shaw-Hwa Lo \\
  Department of Statistics\\
  Columbia University\\
  \texttt{shl5@columbia.edu} \\
   \And
 Yiqiao Yin \\
  Department of Statistics\\
  Columbia University\\
  \texttt{yy2502@columbia.edu} \\
}

\begin{document}
\maketitle
\begin{abstract}
The field of Explainable Artificial Intelligence (XAI) aims to build explainable and interpretable machine learning (or deep learning) methods without sacrificing prediction performance. Convolutional Neural Networks (CNNs) have been successful in making predictions, especially in image classification. However, these famous deep learning models use tens of millions of parameters based on a large number of pre-trained filters which have been repurposed from previous data sets. We propose a novel Interaction-based Convolutional Neural Network (ICNN) that does not make assumptions about the relevance of local information. Instead, we use a model-free Influence Score (I-score) to directly extract the influential information from images to form important variable modules. We demonstrate that the proposed method produces state-of-the-art prediction performance of 99.8\% on a real-world data set classifying COVID-19 Chest X-ray images without sacrificing the explanatory power of the model. This proposed design can efficiently screen COVID-19 patients before human diagnosis, and will be the benchmark for addressing future XAI problems in large-scale data sets.
\end{abstract}


\section{Introduction}
\subsection{AI Systems for COVID-19 Chest X-ray}
The outbreak of novel coronavirus SARS-Cov-2 (previously 2019-nCov, also known as COVID-19) has quickly spread to nearly every country in the world \cite{velavan2020covid} \cite{li2007wang} \cite{wang2021deep}. Since the beginning, the top priority for controlling the spread of COVID-19 has been to monitor suspected cases for appropriate quarantine measures and treatment. Pathogenic laboratory tests are the standard procedure (RT-PCR, collected with the invasive nasal swab most readers will be familiar with), but the accuracy of this test suffers from poor results \cite{velavan2020covid}. Chest CT scans have a high sensitivity for diagnosis of coronavirus disease 2019 (COVID-19) \cite{ai2020correlation} and can be primary tool for the current COVID-19 detection \cite{ai2020correlation}. This innovation, if utilized earlier, could have revolutionized the initial screening procedure of COVID-19 and might prevented many unnecessary deaths. Looking forward, we believe that other related diseases could use similar detection methods to prevent future epidemics and pandemics. Creating these methods now will ensure the development of testing procedures with speed, accuracy, and explainability that will be required to deploy such Artificial Intelligence (AI) systems into future testing environments. AI-centered medical imaging-based deep learning systems have already been developed Convolutional Neural Networks (CNNs) that have shown promising results in feature extraction and learning \cite{wang2021deep}. However, due to increasing complexity of deep CNN models, it is difficult to explain the prediction performance to their human users. To better assist the medical community by providing and deploying Explainable Artificial Intelligence (XAI) systems for analyzing chest scans to save critical time that is necessary for disease control, we call for immediate attention to developing a self-interpretable and explainable Convolutional Neural Network architecture capable of early disease detection.

\subsection{What is XAI?}
\begin{wrapfigure}{r}{0.5\linewidth}
    \centering
    \vspace{-20pt}
    \caption{This diagram is a recreation of the figure in DARPA document (DARPA-BAA-16-53) \cite{DARPA2016} \cite{Rudin2019}. It presents the relationship between learning performance (usually measured by prediction performance) and effectiveness of explanations (also known as explainability). The proposed method in our work aims to take any deep learning method and provide explainability without sacrificing prediction performance. In the diagram, the proposed method is the orange dot on the upper right corner of the relationship plot.}
    \includegraphics[width=0.5\textwidth]{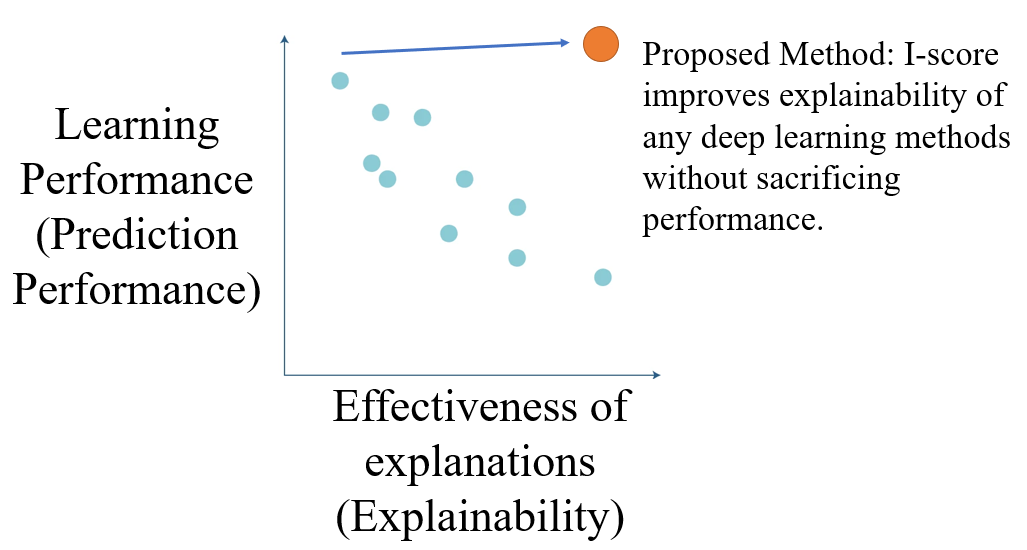}
    \vspace{-20pt}
    \label{fig:accuracy-explanation-tradeoff}
    \vspace{-15pt}
\end{wrapfigure}

In 2016, DARPA initiated the Explainable Artificial Intelligence (XAI) challenge. The goal is to build suites of machine learning algorithms that are interpretable without sacrificing prediction performance (see Figure \ref{fig:accuracy-explanation-tradeoff}). The trade-off between learning performance and the effectiveness of explanation is illustrated in Figure \ref{fig:accuracy-explanation-tradeoff}. The work for this paper delivers a novel interaction-based methodology that produce the power of interpretability and explainability while maintaining state-of-the-art learning performance.

A popular description of interpretability states the essential element for XAI is the ability to explain or to present in understandable terms to a human \cite{DoshiVelez2017TowardsAR}. Another popular version states interpretability as the degree to which a human can understand the cause of a decision \cite{MILLER20191}. Though intuitive, these definitions lack mathematical formality and rigorousness as certain perspectives \cite{Adadi2018}.

In this paper, we regard the core issue of an XAI problem to be ``how features or variables are used to produce the prediction performance''. In other words, the effectiveness of explanations in the DARPA document (see Figure \ref{fig:accuracy-explanation-tradeoff}) is innately a variable set assessment and selection problem. This means the explainability and interpretability of a machine learning algorithm is directly referring to what measures statisticians use to assess how the features or variables affect the final prediction results. In order to establish accountability, responsibility, and transparency in an AI system, one must first address an explainable and interpretable measure to assess the feature importance. We define the following three dimensions, $\mathcal{D}1$, $\mathcal{D}2$, and $\mathcal{D}3$, for a measure to be interpretable and explainable from statistical and variable assessment perspective. In other words, these three dimensions, denoted by $\mathcal{D}1$, $\mathcal{D}2$, and $\mathcal{D}3$, serve as the \textbf{key premises of the definition of an explainable and interpretable measure}.

$\mathcal{D}1$. A quantifiable measure for interpretability and explainability does not need to rely on the knowledge of correct specification of the underlying model. 

$\mathcal{D}2$. Any desirable measure for interpretability and explainability should state clearly how a combination of explanatory variables influence the response variable. Moreover, it can directly compute a score for a set of variables in order to make reasonable comparisons. This means any additional influential variables should raise this measure while any existence of redundant or noisy variables should decrease the magnitude of this measure. 

$\mathcal{D}3$. A measure appropriate for interpretability and explainability needs to state clearly the impact that the influence explanatory variables have on the predictivity of response variable. In other words, this measure should directly speaks for the predictivity (see equation [2] of \cite{lochernoffzhenglo2016}) of the explanatory variables.

\subsection{Problems in Image Classification and Deep CNN}
In the field of image classification, Convolutional Neural Networks (CNNs) are well-known for their high prediction performances in image classification \cite{LeCun89} \cite{Krizhevsky2012} \cite{Simonyan2014} \cite{Huang2016} \cite{He2016} \cite{Chollet2016}. Though CNNs are successful in its application areas, there is no theoretical proof to explain why it performs so well \cite{Aloysius17} especially when using a large size of filters \cite{Khan2020} (See Table \ref{tab:summary-CNN-params}). Another challenge is the underlying assumption of using learned feature-maps due to large filter size \cite{Khan2020} \cite{Krizhevsky2012}. Hence, it is difficult to explain how every parameter contributes to the prediction performance and to deliver explainability that satisfy the three dimensions $(\mathcal{D}1, \mathcal{D}2, \text{ and } \mathcal{D}3)$ defined above.

\begin{table}[H]
    \centering
    \caption{Summary of some well-known CNNs and their number of parameters.}
    \begin{tabular}{lc}
    \toprule
        Name & Number of Parameters \\
        \hline
        LeNet \cite{LeCun89} & 60,000 \\
        AlexNet \cite{Krizhevsky2012} & 60 million \\
        ResNet50 \cite{He2016} & 25 million \\
        DenseNet \cite{Huang2016} & 0.8 - 40 million \\
        VGG16 \cite{Simonyan2014} & 138 million \\
    \bottomrule
    \end{tabular}
    \label{tab:summary-CNN-params}
\end{table}

\subsection{An Interaction-based Convolutional Neural Network (ICNN) to Address XAI Problems}
\begin{figure}
    \centering
    \caption{This executive diagram summarizes the key components of the proposed methods of this paper. We start with the COVID-19 Image Data. With a small rolling window defined, we execute the Backward Dropping Algorithm (BDA) to select the important features within this window. Next, the BDA could select $\{X_1, X_2\}$ as a variable module. Then we can construct a new variable using the technique of Interaction-based Feature Engineer (see the construction of $X^\dagger$ in equation \ref{eq:interaction-based-features-general-form} to appreciate this new design). For more detailed discussion, please see Appendix \ref{fig:main-diagram-appendix}.}
    \includegraphics[width=1\textwidth]{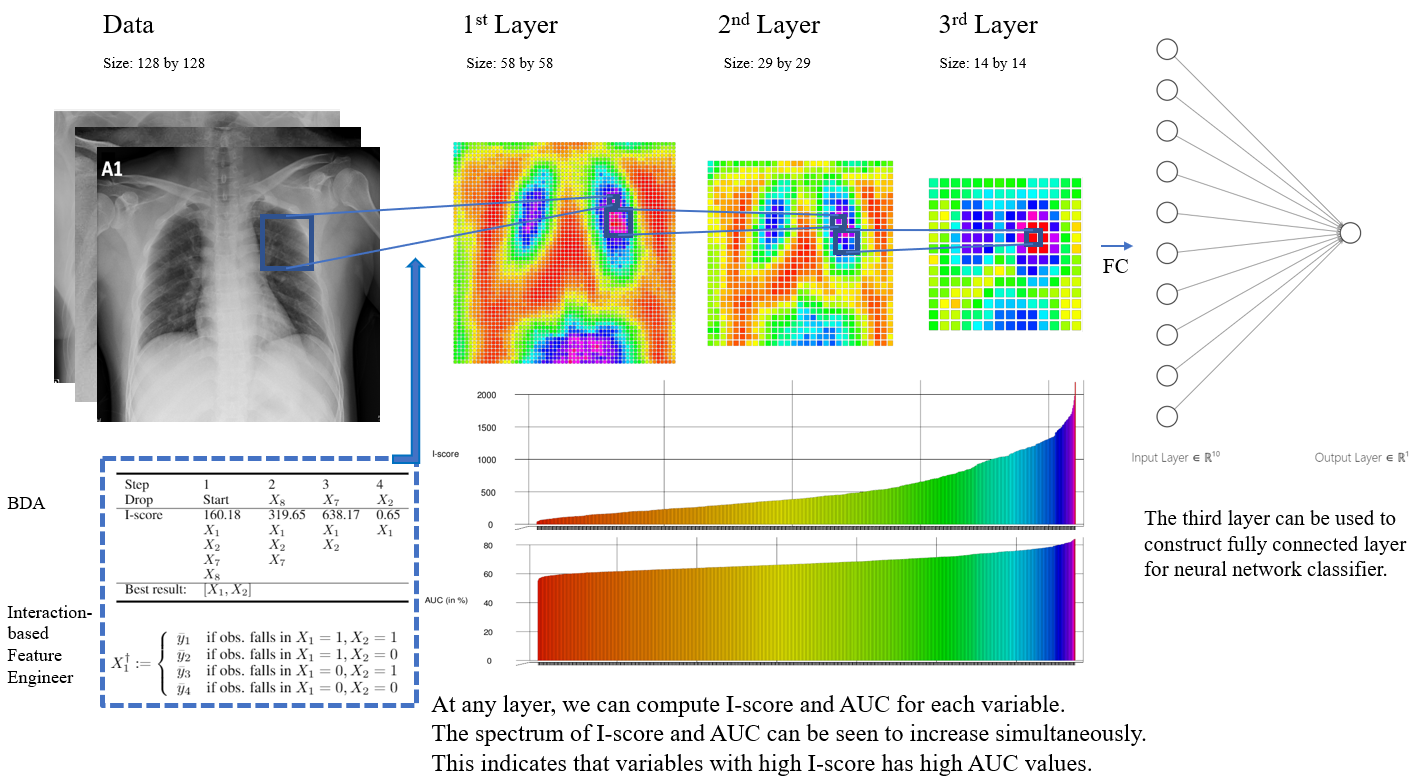}
    \label{fig:main-diagram}
\end{figure}

Chernoff, Lo, and Zheng (2009) \cite{chernoffetal2009} presents a general intensive approach, based on a method pioneered by Lo and Zheng (2002) \cite{lozheng2002} for detecting which, out of many potential explanatory variables, have an influence (impact) on a dependent variable $Y$. This paper presents an interaction-based feature selection methodology incorporating the notion of influence score, I-score, as a major technique to detect the higher-order interactions in complex and large-scale data set. Our work investigates the potential usage of I-score and a novel deep learning framework. The executive diagram can be seen in Figure \ref{fig:main-diagram} which outlines the road-map of the proposed methodology and the architecture of a novel Interaction-based Convolutional Neural Network (ICNN). This novel architecture takes full advantage of I-score and Backward Dropping Algorithm. In other words, it produces convolutional layers that are self-interpretable which provides explainable power to the features of image data at any convolutional layer if this design is implemented. In the following section, we discuss the contribution of our work and why the proposed method satisfies the three dimensions defined in \S1. With all three dimensions satisfied ($\mathcal{D}1$, $\mathcal{D}2$, and $\mathcal{D}3$), the proposed design of an Interaction-based Convolutional Neural Network (ICNN) is the ideal candidate to address XAI problems.

The key novelty of our proposed approach (see Executive Design in Figure \ref{fig:main-diagram}) rests on the collection of features identified by I-score. Based on the contributions described above, the proposed method is model-free and hence checks the first condition $\mathcal{D}1$. It describes a quantifiable measure of how much a combination of explanatory variables impact the outcome variable which allows statisticians to make comparisons and screen for influential features. This phenomenon checks the second dimension $\mathcal{D}2$. In addition, the statistics, I-score, provides a measurement for explanatory variables that is directly associated with the predictivity that the explanatory variables have on outcome variable which satisfies the third condition $\mathcal{D}3$. With all three dimensions ($\mathcal{D}1$, $\mathcal{D}2$, and $\mathcal{D}3$) satisfied, the design of the proposed architecture presents an Interaction-based Convolutional Neural Network (ICNN) that is innately interpretable and explainable. It extracts influential information from the image data and generates explanatory features that directly associate with the predictivity of the data. Because all three dimensions are met, the proposed design of CNN is interpretable and it serves as the touchstone in the field of XAI. For more detailed discussion, please see Appendix \ref{why-satisfied}.

\section{Proposed Method}
\subsection{Influence Score (I-score)}
Assume that we have response variable $Y$ to be binary (taking values 0 and 1) and all explanatory variables to be discrete. Consider the partition $\mathcal{P}_k$ generated by a subset of $k$ explanatory variables $\{X_{b_1}, ..., X_{b_k}\}$. Assume all variables in this subset to be binary. Then we have $2^k$ partition elements; see the first paragraph of Section 3 in (Chernoff et al., 2009 \cite{chernoffetal2009}). Let $n_1(j)$ be the number of observations with $Y = 1$ in partition element $j$. Let $\bar{n}(j) = n_j \times \pi_1$ be the expected number of $Y = 1$ in element $j$. Under the null hypothesis the subset of explanatory variables has no association with $Y$, where $n_j$ is the total number of observations in element $j$ and $\pi_1$ is the proportion of $Y = 1$ observations in the sample. In Lo and Zheng (2002) \cite{lozheng2002}, the influence score is defined as 
\begin{equation}
I(X_{b_1}, ..., X_{b_k}) = \sum_{j \in \mathcal{P}_k} [n_1(j) - \bar{n}_1(j)]^2.
\end{equation}

The statistics $I$ is the summation of squared deviations of frequency of $Y$ from what is expected under the null hypothesis. There are two properties associated with the statistics $I$. First, the measure $I$ is non-parametric which requires no need to specify a model for the joint effect of $\{X_{b_1}, ..., X_{b_k}\}$ on $Y$. This measure $I$ is created to describe the discrepancy between the conditional means of $Y$ on $\{X_{b_1}, ..., X_{b_k}\}$ disregard the form of conditional distribution. Secondly, under the null hypothesis that the subset has no influence on $Y$, the expectation of $I$ remains non-increasing when dropping variables from the subset. The second property makes $I$ fundamentally different from the Pearson's $\chi^2$ statistic whose expectation is dependent on the degrees of freedom and hence on the number of variables selected to define the partition. We can rewrite statistics $I$ in its general form when $Y$ is not necessarily discrete
\begin{equation}\label{eq:iscore}
I = \sum_{j \in \mathcal{P}} n_j^2 (\bar{Y}_j - \bar{Y})^2,
\end{equation}
where $\bar{Y}_j$ is the average of $Y$-observations over the $j$th partition element (local average) and $\bar{Y}$ is the global average. Under the same null hypothesis, it is shown (Chernoff et al., 2009 \cite{chernoffetal2009}) that the normalized $I$, $I/n\sigma^2$ (where $\sigma^2$ is the variance of $Y$), is asymptotically distributed as a weighted sum of independent $\chi^2$ random variables of one degree of freedom each such that the total weight is less than one. It is precisely this property that serves the theoretical foundation for the following algorithm.

\subsection{Backward Dropping Algorithm (BDA)}
The Backward Dropping Algorithm is a greedy algorithm to search for the optimal subsets of variables that maximizes the I-score through step-wise elimination of variables from an initial subset sampled in some way from the variable space. The steps of the algorithm are as follows.

\begin{algorithm}\label{alg:BDA}
\SetAlgoLined
\caption{Procedure of the Backward Dropping Algorithm (BDA)}
\emph{Training Set}: Consider a training set $\{(y_1, x_1), ..., (y_n, x_n)\}$ of $n$ observations, where $x_i = (x_{1i}, ..., x_{pi})\}$ is a $p$-dimensional vector of explanatory variables. The size $p$ can be very large. All explanatory variables are discrete.\;
\emph{Sampling from Variable Space}: Select an initial subset of $k$ explanatory variables $S_b = \{X_{b_1}, ..., X_{b_k}\}$, $b = 1, ..., B$\;
    \emph{Initialization}: Set $l = k$\;
    \While{While $l >= 1$}{
        \emph{Compute Standardized I-score}: calculate $I(S_b) = \frac{1}{n\sigma^2}\sum_{j \in \mathcal{P}_k} n_j^2 (\bar{Y}_j - \bar{Y})^2$. For the rest of the paper, we refer this formula as Influence Measure or Influence Score (I-score). \\
        \emph{Drop Variables}: Tentatively drop each variable in $S_b$ and recalculate the I-score with one variable less. Then drop the one that gives the highest I-score. Call this new subset $S_b'$ which has one variable less than $S_b$. \\
        l = $|S_b'|$ (update $l$ with length of current subset of variables)
    }
\end{algorithm}

\subsection{Interaction-based Convolutional Layer}
\begin{wrapfigure}{r}{0.7\linewidth}
    \centering
    \vspace{-3pt}
    \caption{This is the architecture of Interaction-based Convolutional Neural Network (ICNN). There are two panels. Panel A proposes a single Interaction-based Convolutional Layer while Panel B proposes a deep but similar design.}
    \begin{tabular}{c|c}
        Panel A & Panel B \\
        \includegraphics[width=0.32\textwidth]{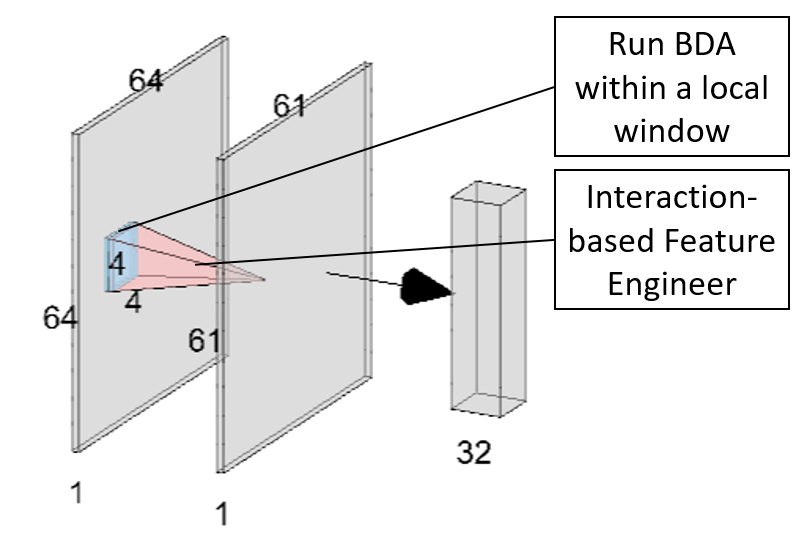} & 
        \includegraphics[width=0.32\textwidth]{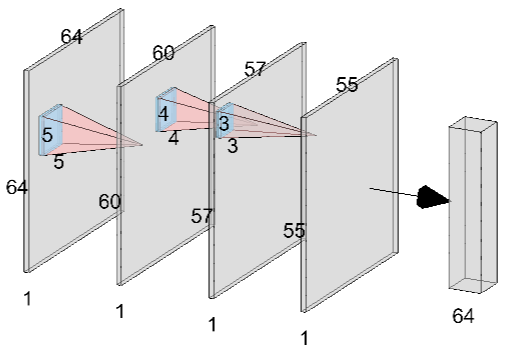} 
    \end{tabular}
    \vspace{-20pt}
    \label{fig:iCNN-architecture}
    \label{fig:deep-iCNN-architecture}
    \vspace{-20pt}
\end{wrapfigure}

This section proposes an Interaction-based Convolutional Neural Network. This action is presented in Figure \ref{fig:iCNN-architecture}. In this diagram, we start with an image that is sized 64 by 64 and suppose we use a window that has shape 4 by 4. Thus, this small window has 16 pixels locally. This set of 16 pixels can be converted into binary variables which hence gives us well defined partition. For example, we can discretize these pixels into black and white. In other words, each pixel takes value 1 or 0 (two levels) and the set of 16 pixels would give us a partition that is sized $2^{16}$. This is the set up for us to run Backward Dropping Algorithm. Based on the definition of the Backward Dropping Algorithm, each round we take turns dropping one variable iteratively. Every time we drop a variable, we compute I-score. We drop the variable at each step such that the I-score is the highest if that variable is dropped. Using this procedure, we are able to select a subset of variables out of the original 16 pixels. A visualization of the real application for this proposed convolutional layer is presented in Row (b) Figure \ref{fig:covid-data-conv-layer-samplewise-plot}.

A deeper but similar architecture can be constructed using the following diagram. In Figure \ref{fig:deep-iCNN-architecture}, we propose a deep Interaction-based Convolutional Neural Network. Instead of building a single Interaction-based Convolutional Layer (ICNN), multiple such layers can be constructed using the same procedure. See Row (b) and Row (c) in Figure \ref{fig:covid-data-conv-layer-samplewise-plot}.

\subsection{Interaction-based Feature Engineer}
This subsection we define a procedure to mechanically engineer an interaction-based feature. 

For an image that is sized $3 \times 3$ (that is, this image has 9 pixels). We can write these pixels as $X_1, X_2, ..., X_9$. Consider a small window of $2 \times 2$ passing from the first row and the first column of this $3 \times 3$ image. This means we start with $\{X_1, X_2, X_4, X_5\}$ within this small 2 by 2 window in this example. We use the Backward Dropping Algorithm to narrow down to $\{X_1, X_2\}$ because we observe this subset of variables deliver us the highest I-score. Since both $X_1$ and $X_2$ are discretized into two partitions, we have partition $\Pi_{\{X_1, X_2\}}$ well defined (see equation \ref{eq:partition-for-interaction-based-feature}). In other words, assume $X_1$ and $X_2$ both only take values 0 and 1. Then the partition $\Pi_{\{X_1, X_2\}}$ is defined as the following
\begin{equation}\label{eq:partition-for-interaction-based-feature}
    \Pi := 
    \left\{
    \begin{array}{lcl}
        \pi_1 & \text{ if } X_1 = 1, X_2 = 1 \\
        \pi_2 & \text{ if } X_1 = 1, X_2 = 0 \\
        \pi_3 & \text{ if } X_1 = 0, X_2 = 1 \\
        \pi_4 & \text{ if } X_1 = 0, X_2 = 0 \\
    \end{array}
    \right.
\end{equation}
In this case, each partition $\pi_j$ while $j \in \{1, 2, 3, 4\}$ we can compute the local mean of response variable $Y$ from observations in training set. Hence, a new interaction-based feature can be constructed using $X^{\dagger} := \bar{y}_j \text{ while } \bar{y}_j \text{ is the local mean of } Y \text{ based on the } j^\text{th} \text{ partition } \in \Pi$. In general situation, let us consider an input matrix with size $s_\text{in}$ by $s_\text{in}$. Suppose window size is $w \times w$ and the stride level to be $l$ (notice in the above example $w = 2$ and $l = 1$). By, $s_\text{out} = \floor{\frac{s_\text{in} - w}{l} + 1}$, we can compute output matrix with size $s_\text{out}$ by $s_\text{out}$ (which coincide with the $X^\dagger$ matrix). In this case, we define running index $b$ to take values $\{1, 2, 3, ... s_\text{out} \times s_\text{out}\}$. In the example in previous paragraph, $b$ can take value $\{1, 2, 3, 4\}$ because $s_\text{out} = \floor{\frac{s_\text{in} - w}{l} + 1} = \floor{(3 - 2)/1 + 1} = 2$ thus $s_\text{out} \times s_\text{out} = 4$. For each round $b$ of Backward Dropping Algorithm ($b$ takes value from $\{1,2,3,4\}$ in this example), we can construct a new variable module that takes formula \ref{eq:interaction-based-features-general-form}, defined below,
\begin{equation}\label{eq:interaction-based-features-general-form}
    \begin{array}{rcl}
        X^{\dagger}_b &:=&
        \bar{y}_j \text{ while } \bar{y}_j \text{ is the local mean of } Y \\
        && \text{ based on the } j^\text{th} \text{ partition } \in \Pi
    \end{array}
\end{equation} while $X_b^\dagger$ is defined as $\bar{y}_j$ using observations in training set while $j$ indicates the $j^\text{th}$ partition of $\Pi$ that is formed by the variables selected from the $b$ round of Backward Dropping Algorithm. Hence, the relationship from a 3-by-3 input matrix into a 2-by-2 output matrix can be visualized using the following diagram
\begin{equation*}
\text{input:}
\begin{bmatrix}
    X_1 & X_2 & X_3 \\
    X_4 & X_5 & X_6 \\
    X_7 & X_8 & X_9 \\
\end{bmatrix}_\text{3 by 3}
\rightarrow
\text{output:}
\begin{bmatrix}
    X_1^\dagger & X_2^\dagger \\
    X_3^\dagger & X_4^\dagger \\
\end{bmatrix}_\text{2 by 2}
\end{equation*}
while the input matrix has size 3 by 3 and the output matrix has size 2 by 2. In the output matrix, the $X^\dagger$'s are defined using equation \ref{eq:interaction-based-features-general-form} with observations in training set.

\section{Application}
\subsection{Data}
The data is sourced from the work by \cite{Minaee2020}. For detailed discussion of the source of data, please see Appendix \ref{data-source}. We present a brief table for the data in Table \ref{tab:covid-data-size-summary}. 

\begin{wrapfigure}{r}{0.7\linewidth}
    \centering
    \vspace{-18pt}
    \caption{The figure presents two panels. Panel A exhibits 16 randomly sampled images from class ``COVID''. Panel B exhibits 16 randomly sampled images from class ``non-COVID''. We observe that the images for the COVID class appear more cloudy and unclear than the images for the non-COVID class. This is because in the X-ray images for COVID class there are substance that are not air. These substances can be liquid, germs, or inflammatory fluid which causes the images to have cloudy, unclear, and shady areas.}
    \begin{tabular}{ll}
    Panel A & Panel B \\
    \includegraphics[width=0.33\textwidth]{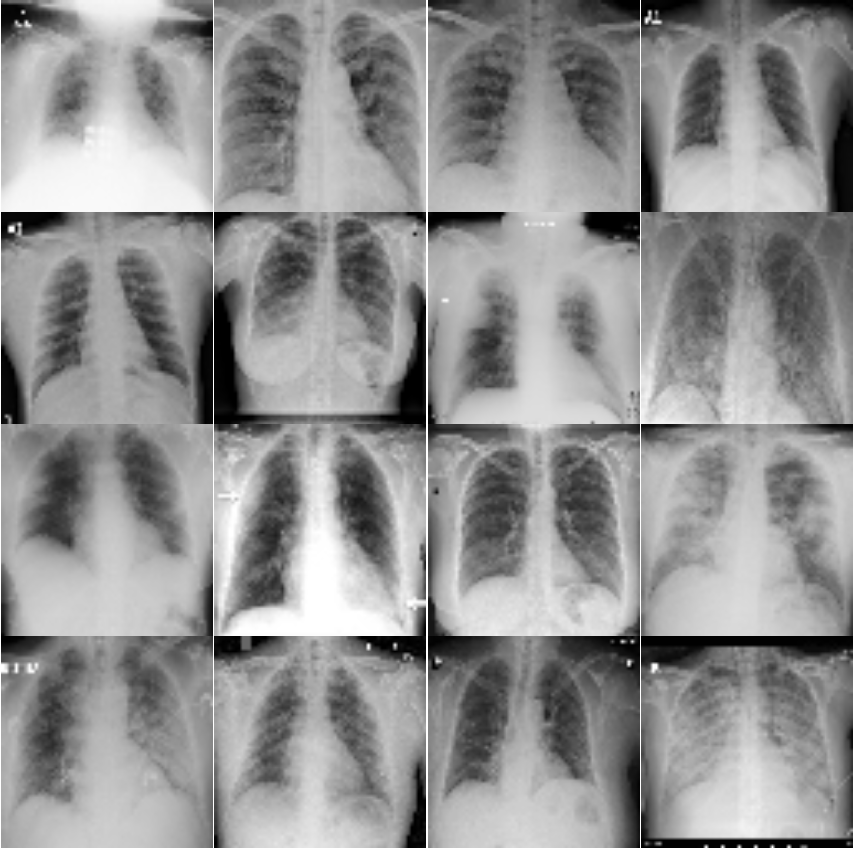} & \includegraphics[width=0.33\textwidth]{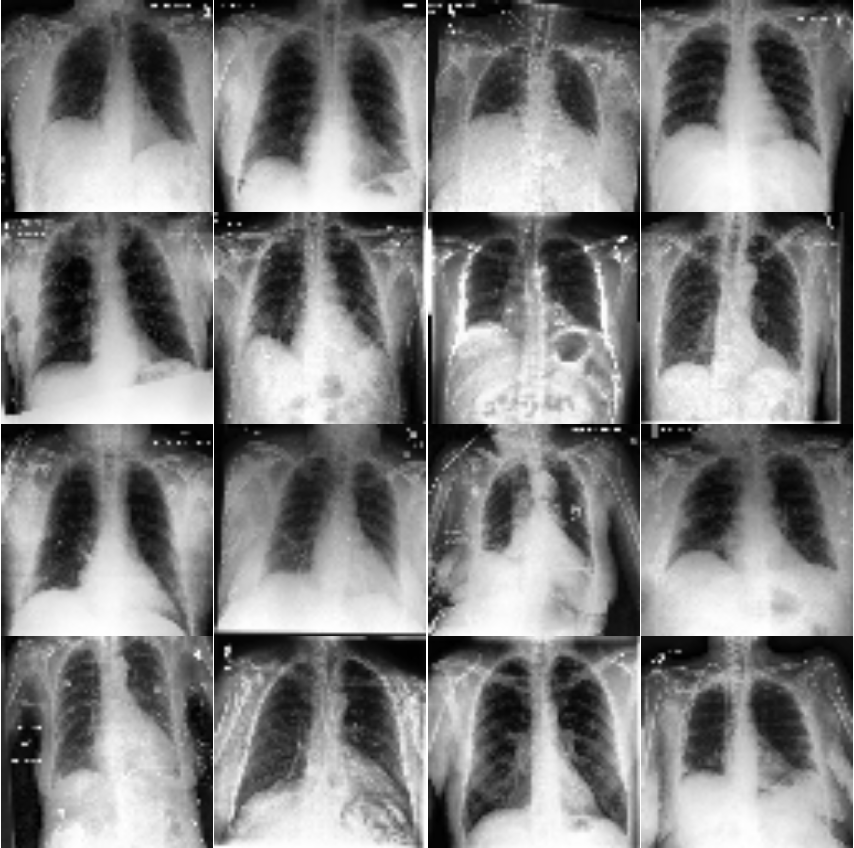} \\
    \end{tabular}
    \vspace{-20pt}
    \label{fig:covid-data-classcovid-vs-classnoncovid}
    \vspace{-20pt}
\end{wrapfigure}

From observation of Figure \ref{fig:covid-data-classcovid-vs-classnoncovid}, it is visible for humans that the chest areas are clear in the healthy chest X-ray images. However, the images for COVID cases are not quite clear. This is an indication that there are inflammatory cells or other related body fluids filled in the chests. Instead of air which shows up on the pictures to be clear area, these areas in COVID cases tend to be cloudy and unclear.

\begin{table}
    \centering
    \caption{This table summarizes the dimension of the data. We download the COVID data from the work by \cite{Minaee2020}. This totalled 576 COVID images and 2,000 non-COVID images. First, we split test set from the total images. The test consists of 60 COVID cases and 60 non-COVID cases. Next, we are left with 516 images for COVID class and 1,940 images for non-COVID class. This is in-sample set which we use for training and validating (short of Tr. and Val. below). For in-sample set, we upsample the images by adding noises drawn from normal distribution. This gives us 5,000 COVID images and 5,000 non-COVID images for training and testing. The out-of-sample test set has 120 observations and this is only used in the end to check the learning performance.}
    \begin{tabular}{lccccc}
        \toprule
        Data & COVID & Non-COVID \\
        \hline
        Total Data Downloaded from \cite{Minaee2020} & 576 & 2,000 \\
        Out-of-Sample: Test & 60 & 60 \\
        In-Sample: Tr. and Val. & 516 & 1,940 \\
        In-Sample: Tr. and Val. & 5,000 & 5,000 \\
        (upsampled) \\
        \bottomrule
    \end{tabular}
    \label{tab:covid-data-size-summary}
\end{table}

\subsection{Performance: Proposed Models}
\begin{figure}
    \centering
    \caption{This figure presents the AUC path for all six models in the proposed work. These models are listed in Table \ref{tab:covid-data-experimental-results} with detailed information including the parameters from each layer and the out-of-sample prediction performance.}
    \begin{tabular}{cc}
        \includegraphics[width=0.5\textwidth]{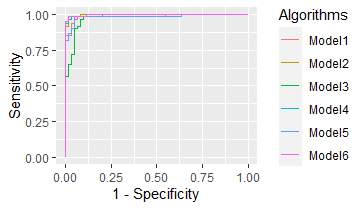} \\
    \end{tabular}
    \label{fig:covid-data-multiple-auc-change-window-size-stride-level}
\end{figure}

This subsection we present the experimental results. A brief summary of comparison of conventional methods literature and the proposed method is presented in Table \ref{tab:covid-data-benchmark-results}. A detailed report of the proposed methodology is presented in Table \ref{tab:covid-data-experimental-results}. We also plot the AUC values for the proposed models in Figure \ref{fig:covid-data-multiple-auc-change-window-size-stride-level}. 

\begin{table}
    \centering
    \caption{The table presents experimental results of COVID-19 data set from literature. A number of different ultra-deep CNNs are used to classify COVID patients from non-COVID people. The average number of parameters of the ultra-deep CNNs can exceed 25 million parameters with top AUC to be at 99.2\%. The proposed methods have average number of parameters to be less than 100k with top AUC of 99.8\%. This is a 99\% reduction on number of parameters without sacrificing the prediction performance.}
    \begin{tabular}{llc}
        \toprule
        Previous Work & Number of Param. & AUC \\
        \hline
        DenseNet161 \cite{Minaee2020} & 0.8 - 40 mil. param. & 97.6\% \\
        ResNet18 \cite{Minaee2020} & 11 mil. param. & 98.9\% \\
        ResNet50 \cite{Minaee2020} & 25 mil. param. & 99.0\% \\
        SqueezeNet \cite{Minaee2020} & $\sim$ 1.2 mil. param.* & 99.2\% \\
        Average & $> 25$ mil. & 97\% - 99.2\% \\
        \hline
        Proposed & average 100k param. & 98.3\% - 99.8\% \\
        & (a 99\% reduction \\
        & on num. of param.) \\
        \bottomrule
    \end{tabular}
    \label{tab:covid-data-benchmark-results}
\end{table}

The report in Table \ref{tab:covid-data-benchmark-results} shows results of previous work on this data set. Minaee et al. (2020) \cite{Minaee2020} used a 50-layer Convolutional Neural Network, ResNet50, and delivered a value of AUC at 99\%. Their work also proposed SqueezeNet that delivered an even higher performance at 99.2\% \cite{Minaee2020}. This near-perfect performance is largely due to the design of the convolutional layers and fine tuning. Their work show that modern day AI technology such as deep Convolutional Neural Network can provide initial screening of COVID-19 diseased patients with a single scan of an image which could provide reduce workload for radiologists in practice. However, the number of parameters still far exceeds what humans can interpret. Moreover, it is unclear how these conventional methodologies can satisfy the three dimensions ($\mathcal{D}1$, $\mathcal{D}2$, and $\mathcal{D}3$) of the definition of interpretable measures introduced in \S1 of this paper.

We read the experimental results of the proposed model in the following order. As stated in section \S3.3, the proposed architecture can be designed as deep or wide as the user desires. All models start with input images with dimensions 128 by 128. In other words, the input data has $128 \times 128 = 16,384$ pixels. For simplicity of notation, let us denote $\triangle$ to be the set of parameters of starting point of 6, window size of 2 by 2, and a stride level of 2. Let us further denote $\square$ to be the collection of parameters of starting point of 1, window size of 2 by 2, and a stride level of 2.

For the training of the proposed model and the tuning parameters, please see Appendix \ref{model-training}, \ref{model-parameters}, and \ref{model-metrics}.

\textbf{Model 1.} This model starts with input images that are sized 128 by 128. Using the parameter in set $\triangle$, we create the first Interaction-based Convolutional Layer (namely 1st Conv. in Table \ref{tab:covid-data-experimental-results}). This new matrix has dimension $\floor{(128-6-2+1)/2+1} \times \floor{(128-6-2+1)/2+1} = 61 \times 61 = 3,721$. These 3,721 variables are used directly to create output layer with 2 units (assuming using softmax as loss function). Therefore, the total number of parameters for the network architecture $3,721 \times 2 = 7,442$ parameters. The test set performance, measured by AUC, is 98.5\% for Model 1.

\textbf{Model 2 - 6.} We also provided more updated versions of Model 1. Please see appendix. \ref{updated-models}

\begin{table}
    \centering
    \caption{The table presents the summary statistics of the design of the proposed network: Interaction-based Convolutional Neural Network (ICNN). There are models 1-6. Each model can take one or two Interaction-based Convolutional Layer (i.e. 1st Conv. or 2nd Conv.). The design of the model can directly go from Interaction-based Convolutional Layer to Output Layer. For example, Model 1 and Model 3 go directly from convolutional layer to output layer, i.e. no hidden layer.}
    \resizebox{\textwidth}{!}{%
    \begin{tabular}{lllllll}
        \toprule
        \textbf{Proposed Work} & 1st Conv. & 2nd Conv. & Hidden & Output Layer & Num. of Param. & AUC \\
        \hline
        Model 1 & $\triangle$ & None & None & 2 & 7,442 & 98.5\% \\
        Model 2 & $\triangle$ & None & 1L(64 units) & 2 & 238,272 & 99.7\% \\
        Model 3 & $\triangle$ & $\square$ & None & 2 & 1,800 & 97.0\% \\
        Model 4 & $\triangle$ & $\square$ & 1L(64 units) & 2 & 57,728 & 99.6\% \\
        Model 5 & $\triangle + \square$ & None & None & 2 & 9,242 & 98.3\% \\
        Model 6 & $\triangle + \square$ & None & 1L(64 units) & 2 & 295,872 & 99.8\% \\
        \hline
        Remark & $\triangle$: & $\square$: \\
            & Starting Point = 6 & Starting Point = 1 \\
            & Window Size: 2 by 2, & Window Size: 2 by 2, \\
            & Stride = 2, & Stride = 2, \\
            & Output: 61 by 61 & Output: 30 by 30 \\
        \bottomrule
    \end{tabular} }
    \label{tab:covid-data-experimental-results}
\end{table}

\subsection{Visualization: Images and Convolutional Layer}
\begin{figure}
    \centering
    \caption{This figure presents visualization summary for 10 randomly sampled images from COVID class and non-COVID class (each has 10). Panel A is for COVID and Panel B is non-COVID. Each panel samples 10 images from the data. Each of the 10 images is then used to feed into proposed architecture and the visualization below presents these images after each proposed Interaction-based Convolutional Layer. For detailed discussion, please see Appendix \ref{fig:covid-data-conv-layer-samplewise-plot-appendix}.}
    \resizebox{\textwidth}{!}{%
    \begin{tabular}{c|l|l}
        \toprule
        & Panel A: & Panel B \\
        & True Label: COVID & True Label: Non-COVID \\
        \hline 
        & Input Images: 128 by 128 & Input Images: 128 by 128 \\
        & (Randomly select 10 samples) & (Randomly select 10 samples) \\
        Row (a) & \includegraphics[width=.47\textwidth]{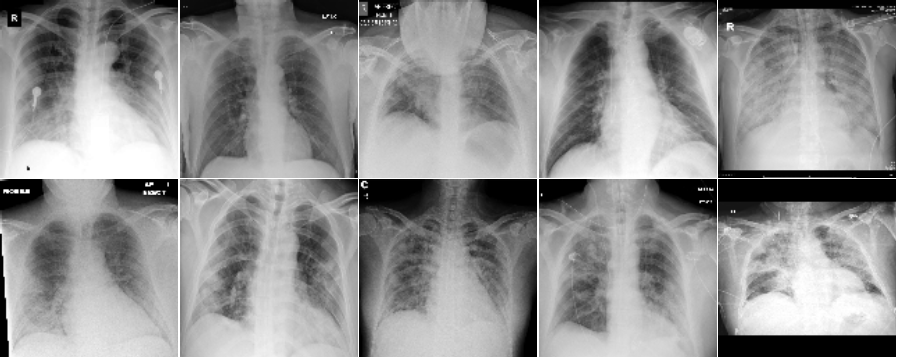} & \includegraphics[width=.47\textwidth]{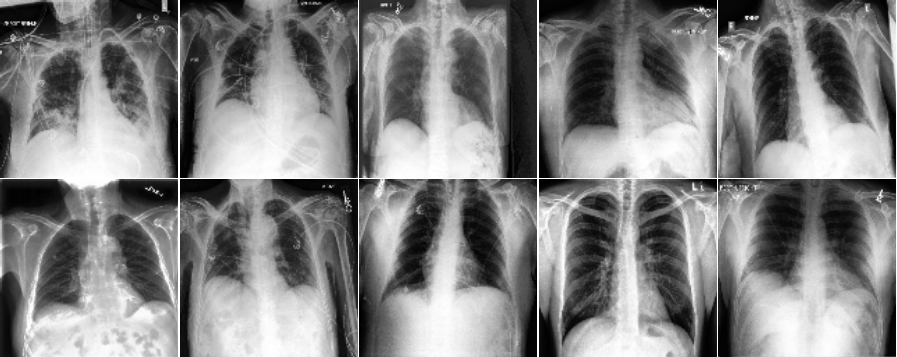} \\
        \hline
        & 1st Conv. Layer: 61 by 61 & 1st Conv. Layer: 61 by 61 \\
        & (Starting Point = 6, Window 2 by 2, Stride = 2) & (Starting Point = 6, Window 2 by 2, Stride = 2) \\
        & Remark: $61 \times 61 = 3,721$ variables & Remark: $61 \times 61 = 3,721$ variables \\
        & Same 10 images above with 3,721 variables & Same 10 images above with 3,721 variables \\
        & Labels predicted using Model 1 & Labels predicted using Model 1 \\
        Row (b) & \includegraphics[width=.47\textwidth]{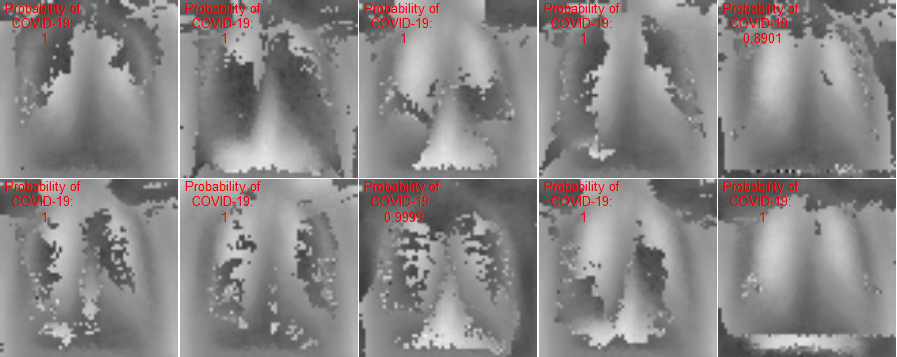} & 
        \includegraphics[width=.47\textwidth]{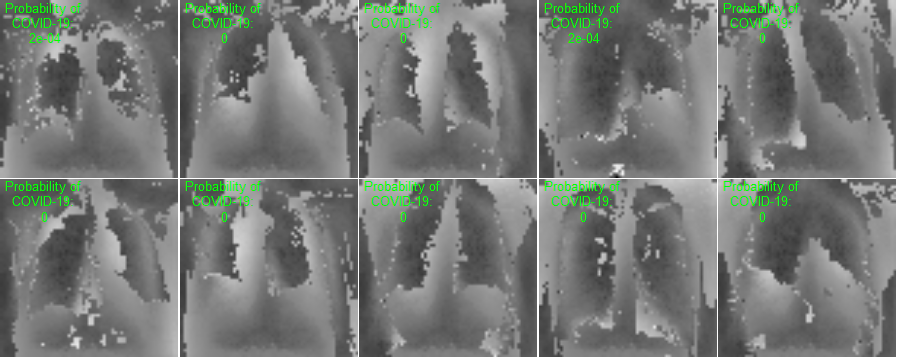} \\
        \hline
        & 2nd Conv. Layer: 30 by 30 & 2nd Conv. Layer: 30 by 30 \\
        & (Starting Point = 6, Window 2 by 2, Stride = 2) & (Starting Point = 6, Window 2 by 2, Stride = 2) \\
        & Remark: $30 \times 30 = 900$ variables & $30 \times 30 = 900$ variables \\
        & Same 10 images above with 900 variables & Same 10 images above with 900 variables \\
        & Labels predicted using Model 4 & Labels predicted using Model 4 \\
        Row (c) & \includegraphics[width=.47\textwidth]{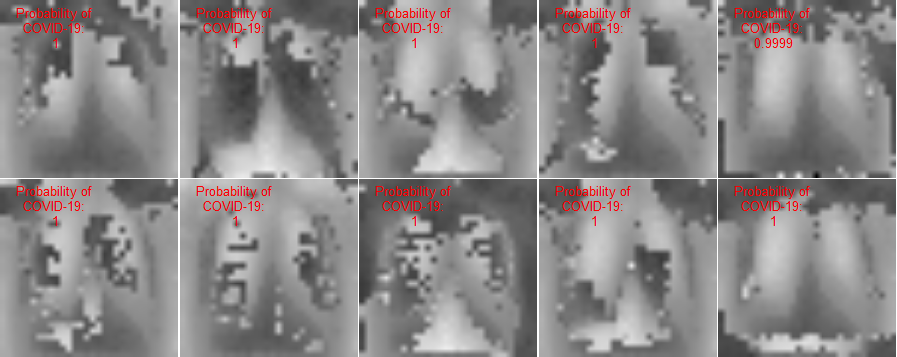} & 
        \includegraphics[width=.47\textwidth]{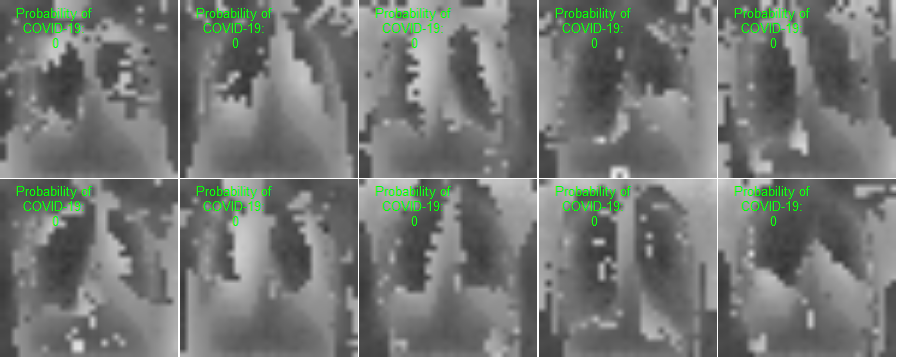} \\
        \bottomrule
    \end{tabular} }
    \label{fig:covid-data-conv-layer-samplewise-plot}
    \vspace{-22pt}
\end{figure}

This section presents visualization of the proposed architecture. These visualizations are presented in Figure \ref{fig:covid-data-conv-layer-samplewise-plot}. Unlike Figure \ref{fig:main-diagram} that is an executive summary with each position representing many samples, these visualizations in Figure \ref{fig:covid-data-conv-layer-samplewise-plot} are sample-wise plots. In other words, the 10 original images that are sized 128 by 128 in Panel A and Panel B are the same samples in the second row, 1st Conv. Layer, and the third row, 2nd Conv. Layer. 

\textbf{Visualization Interpretation} The plot in Figure \ref{fig:covid-data-conv-layer-samplewise-plot} of the original images for COVID-19 patients has grey and cloudy textures in chest area. Because an X-ray picture is at its brightest when most of the light beams emitted are bounced back from the object, we can observe bones to be the color ``white'' while the margin to be completely ``black''. For muscle and organs inside human body, X-ray beams that are emitted can only partially be collected and this causes the greyscale on the X-ray images in chest area. For COVID-19 patients, there are grey and shaded area in the chest X-ray pictures. This is due to the inflammatory fluid when patients exhibit pneumonia-like symptoms. The fluid inside chest area is a consequence of human immune system fighting against outside diseases. This shaded (as seen in Panel A of Figure \ref{fig:covid-data-conv-layer-samplewise-plot}) prevents us from observing the clear location of lungs. This is different in Panel B where the lung areas are dark and almost black, because a healthy lung is filled with air (i.e. normal cases and X-ray image presents color black). The black and white contrast in the two panels is directly related to how much inflammatory fluid there is in human lungs. This contrast translates to greyscale on pictures and it is directly related with COVID cases and non-COVID cases (i.e. response variable $Y$). The same contrast can be seen using the new variables (these are $X^\dagger$'s based on equation \ref{eq:interaction-based-features-general-form}) in the 1st Conv. Layer (sized 61 by 61). For COVID-19 patients, the lung area is cloudy and unclear while the healthy cases it is clearly visible. \textbf{This is not a surprising coincidence because the proposed new variable modules, $X^\dagger$'s, are engineered using equation \ref{eq:interaction-based-features-general-form} which relies on the response variable $\bar{y}_j$ in training set. The images sized 61 by 61 from the proposed algorithm is a direct translation of not only the original pixels but also response variable. In other words, this visualization presents how I-score sees image data.}

\section{Conclusions}
\textbf{Explainable AI System for Early COVID-19 Screening} As the most important contribution of this paper, an Explainable Artificial Intelligence (XAI) system is proposed to assist radiologists for the initial screening of COVID-19 and other related diseases using chest X-ray images for treatment and disease control with accountability, responsibility, and transparency to human users and patients.

\textbf{A Heuristic and Theoretical Framework of XAI.} This paper introduces a heuristic and theoretical framework for addressing the XAI problems in large-scale and high-dimensional data sets. We provide three dimensions ($\mathcal{D}1$, $\mathcal{D}2$, $\mathcal{D}3$) as necessary conditions and premises required for a measure to be regarded as explainable and interpretable.

\textbf{An Interaction-based Convolutional Neural Network (ICNN).} To address the XAI problems heuristically described above, this paper introduced a novel design of an explainable and self-interpretable Interaction-based Convolutional Neural Network (ICNN) to contribute to the major issues about explainability, interpretability, transparency, and trustworthiness in black-box algorithms. We introduce and implement a non-parametric and interaction-based feature selection methodology and use this as replacement of pre-defined filters that are widely used in ultra-deep CNNs. Under this paradigm, we present an Interaction-based Convolutional Neural Network (ICNN) that extract important features. We believe that the proposed design can be adapted in any type of CNN. Thus, any CNN architecture that adapts the proposed technology can be regarded as Interaction-based Convolutional Neural Network (ICNN or Interaction-based Network). We encourage both the statistics and computer science communities to further explore this area to deliver more transparency, trustworthiness, and accountability to deep learning algorithms and to build a world with truly Responsible A.I.. \label{limitations}

\subsection*{Acknowledgment}
We would like to dedicate this to H. Chernoff, a well-known statistician and a mathematician worldwide, in honor of his 98th birthday and his contributions in Influence Score (I-score) and the Backward Dropping Algorithm (BDA). We are particularly fortunate in receiving many useful comments from him. Moreover, we are very grateful for his guidance on how I-score plays a fundamental role that measures the potential ability to use a small group of explanatory variables for classification which leads to much broader impact in fields of pattern recognition, computer vision, and representation learning.

\subsection*{Disclosure of Funding}
This research is supported by National Science Foundation BIGDATA IIS 1741191.

\bibliographystyle{unsrt}  


\newpage
\section{Appendix}

\subsection{Why the Proposed Methodology Satisfies XAI Dimensions?} \label{why-satisfied}
The illustration of the proposed Interaction-based Convolutional Nerual Network (ICNN) is presented in Figure \ref{fig:main-diagram}. This executive diagram walks reader through a step-by-step procedure how the proposed network architecture is designed and why it satisfies the three dimensions ($\mathcal{D}1$, $\mathcal{D}2$, and $\mathcal{D}3$) in the definition of interpretability and hence can be the benchmark address XAI problems.

\textbf{Proposed Architecture}. The executive diagram for the proposed architecture is presented in Figure \ref{fig:main-diagram}. First, the architecture starts with image data that consists of X-ray pictures sized 128 by 128 (see detailed discussion of COVID-19 data set in \S4). The architecture proposes using a rolling window with size 2 by 2 (we use a 2 by 2 window for simplicity and larger sizes can be applicable as well in practice depending on the data). Since the window size is 2 by 2, this means there would be four variables every time the window rests on a certain location of the image. Within this subset of variables, we execute the proposed method of Backward Dropping Algorithm (BDA). This procedure finely selects a subset of variables that is highly predictive by omitting the noisy variables in this small neighborhood on the image. Next, the selected variables (which can be any subset of the original four) go through a proposed procedure called Interaction-based Feature Engineer (see \ref{eq:interaction-based-features-general-form} for definition). The procedure of BDA is illustrated in the bottom left corner of the Figure \ref{fig:main-diagram} (we use a 2 by 2 window for demonstration purpose). In addition, we set the starting point to be 12 and this means we start from the pixel at the 12th row and the 12th column. From data (size of 128 by 128) to the 1st layer (58 by 58), this procedure gives us a new feature matrix with size $\floor{(128-12-2+1)/2 + 1} = 58$ on both edges which means the new feature matrix has 58 by 58 variables (the formula is presented in equation \ref{eq:new-dim}). This feature matrix constitutes the first Interaction-based Convolutional Layer. We can then use the same methodology to construct the second and the third Interaction-based Convolutional Layer. The third Interaction-based Convolutional Layer can be used as input layer for a neural network classifier. Each layer we can compute the proposed measure I-score and the AUC value (see \S4.4 for detailed discussion of AUC values) for each variable (assuming using this variable as a predictor when computing the AUC value). The paths of I-score and AUC values have parallel behavior and this can be seen in the color palette of the spectrum in Figure \ref{fig:main-diagram}.

\textbf{Why Proposed I-score Satisfies XAI Dimensions}. The design of the proposed architecture in Figure \ref{fig:main-diagram} mainly focuses on using I-score and Backward Dropping Algorithm to extract and engineer features from the original images. The proposed measure I-score is non-parametric (see \S3.1 for definition of this measure). This means the impact of how explanatory variables affect response variable measured by I-score does not rely on the knowledge of the correct specification of the underlying model. \textbf{In other words, the computation of I-score does not rely on any model fitting procedure.} This characteristic satisfies the first dimension, $\mathcal{D}1$, defined in \S1 about interpretable measures. 

Next, the proposed architecture is transparent at disclosing to its human users what location of the image is important at making decisions about prediction and what location is noisy. In every Interaction-based Convolutional Layer, we can compute the proposed measure I-score for any single variable. We can also compute I-score and finely select predictive variables from any combination of variables. The larger the I-score values the more important the variables are at making predictions. With simple visualization presented in Figure \ref{fig:main-diagram}, we are able to use a spectrum of different colors to illustrate this phenomenon. We can code high I-score values to be one side of the color spectrum and the low I-score values to be the opposite side. \textbf{The areas that are informative have high I-score values and have very different color than areas that are noisy. This characteristics of I-score allows the statistician to make comparisons and variable selection assessment.} Therefore, it satisfies the second dimension, $\mathcal{D}2$, defined in \S1. 

Third, the proposed architecture has a direct association with the predictivity (see equation[2] of \cite{lochernoffzhenglo2016}) of a variable set. This means that the important features screened by I-score provide the statistician how much impact this variable set has on the response variable. In addition, it is beneficial to be able to compute I-score for any variable in any Interaction-based Convolution Layer which implies that the association with the predictivity is well stated each step in the architecture. \textbf{This can be visualized using the AUC values (see \S4.4 for detailed discussion of AUC values) that are coded onto the same spectrum location of I-score values. Moreover, we observe the path of I-score and AUC values to have parallel behavior. This implies that the values of I-score and AUC move up and down together which means for each variable with high I-score measure would definitely have high AUC values and vice versa.} This interesting yet powerful phenomenon allows this architecture to satisfy the third dimension, $\mathcal{D}3$, stated in the definition of an interpretable measure which is novel in the literature. 

As a summary, we have identified that the proposed Interaction-based Convolutional Neural Network (ICNN) relies on I-score which is a measure satisfies the three dimensions of explainable and interpretable measure. We regard the proposed ICNN to be an explainable and interpretable deep learning algorithm. Thus far, we have not been able to identify other methods and procedures that satisfies all three dimensions defined in \S1.

\subsection{Model Training} \label{model-training}
For the neural network classifier adapted in this paper, we use a standard neural network architecture with some basic designs (see the diagram after equation \ref{eq:activation-function}). This procedure has input variables (known as input layer), an optional hidden layer, and also output layer. The input layer are sets of variables ready to be fed into the machine to build classifier. The hidden layer can consist of any number of units and it is an optional design to deepen the network architecture. The output layer is constructed in order to compare with the output variable. This comparison can be quantified by a loss function. The path from input layer to output layer completes the procedure of forward propagation. Based on the loss function, we are able to compute the gradient which allows us to conduct optimize the weights of the architecture backwards by using gradient descent (or some upgraded version of gradient descent). This completes backward propagation. \textbf{The training of a neural network model is based on many iterations of forward and backward propagation.}

\textbf{Forward Propagation.} To illustrate the procedure of model training. Let us consider a set of input variables to be $\{X^\dagger_1, X^\dagger_2, X^\dagger_3\}$. In the proposed work, this is referring to the variable modules, also notated as $X^\dagger$'s, that we created using Interaction-based Feature Engineer (see equation \ref{eq:interaction-based-features-general-form}). For this discussion, we define a set of weights $\{w_1, w_2, w_3\}$ to construct a linear transformation. The symbol $\Sigma$ below in the following diagram represents this linear transformation that takes the form $X^\dagger_1w_1 + X^\dagger_2w_2 + X^\dagger_3w_3$. For simplicity of notation, we write $\Sigma = \sum_{j=1}^3 w_j X^\dagger_j$ in short. Then we denote $a(\cdot)$ as an activation function. We choose sigmoid to be this activation function $a(\cdot)$. This means we have output $\hat{y}$ to be defined as $a(\Sigma)$. In other words, let us write the following
\begin{equation}\label{eq:activation-function}
    \hat{y} := a(\Sigma) = a(\sum_{j=1}^3 w_j X^\dagger_j) = 1/(1+\exp(-(\sum_{j=1}^3 w_j X^\dagger_j)))
\end{equation}
The general form (assuming there are $p$ variable modules) is expressed below
\begin{equation}\label{eq:activation-function-general}
    \hat{y} := a(\Sigma) = a(\sum_{j=1}^p w_j X^\dagger_j) = 1/(1+\exp(-(\sum_{j=1}^p w_j X^\dagger_j)))
\end{equation}

\textbf{Architecture.} This architecture of neural network is presented below. For simplicity of drawing this picture, we assume there are 3 input variable modules, $\{X^\dagger_1, X^\dagger_2, X^\dagger_3\}$. In practice, the number of variable modules (the total number of $X^\dagger$'s) depends on image data dimensions, window size, stride level, and starting point (please see \S4.3 and equation \ref{eq:new-dim} for the exact calculation). 

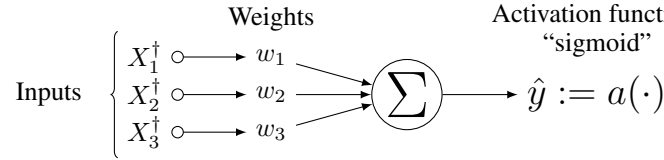
\begin{figure}[H]
    \centering
    \caption{The architecture below presents a feed forward neural network with three interaction-based input variables. The input variables are $\{X^\dagger_1, X^\dagger_2, X^\dagger_3\}$ which are variable modules created using equation \ref{eq:interaction-based-features-general-form}.}
    \begin{tikzpicture}
        [
            init/.style={ 
                 draw, 
                 circle, 
                 inner sep=1pt,
                 font=\Huge,
                 join = by -latex
            },
            squa/.style={ 
                font=\Large,
                join = by -latex
            }
        ]
        \begin{scope}[start chain=1]
            \node[on chain=1] at (0,.5cm)  (x1) {$X^\dagger_1$};
            \node[on chain=1,label=above:Weights,join=by o-latex] (w1) {$w_1$};
        \end{scope}
        \begin{scope}[start chain=2]
            \node[on chain=2] (x2) {$X^\dagger_2$};
            \node[on chain=2,join=by o-latex] {$w_2$};
            \node[on chain=2,init] (sigma) {$\displaystyle\Sigma$};
            \node[on chain=2,squa,label=above:{\parbox{3cm}{\centering Activation function:\\ ``sigmoid''}}] {$\hat{y} := a(\cdot)$};
        \end{scope}
        \begin{scope}[start chain=3]
            \node[on chain=3] at (0,-.5cm) 
            (x3) {$X^\dagger_3$};
            \node[on chain=3,join=by o-latex]
            (w3) {$w_3$};
        \end{scope}
        \draw[-latex] (w1) -- (sigma);
        \draw[-latex] (w3) -- (sigma);
        \draw[decorate,decoration={brace,mirror}] (x1.north west) -- node[left=10pt] {Inputs} (x3.south west);
    \end{tikzpicture}
    \label{fig:nn-architecture}
\end{figure}

For the loss function, we used the binary cross-entropy loss function. This loss function is designed to minimize the distance between a target probability distribution $P$ and an estimated target distribution $Q$ when the task is a two-class classification problem. The cross-entropy loss function is defined as the following
\begin{equation}\label{eq:cross-entropy}
    \begin{array}{l}
        \mathcal{L}(y_i, \hat{y}_i) = -\frac{1}{n} \sum_{i=1}^n y_i \log(\hat{y}_i) + (1 - y_i) \log(1 - \hat{y}_i) \\
    \end{array}
\end{equation}
where $y_i$ is the ground truth of response variable for the $i^\text{th}$ observation and $\hat{y}_i$ is the predicted value of response variable for the $i^\text{th}$ observation. The linear transformation, non-linear transformation, and the computation of the loss function completes the forward propagation.

\textbf{Backward Propagation.} To search for the optimal weights, we use an optimizer algorithm called RMSprop (short for Root Mean Square Propagation, a named suggested by Geoffrey Hinton). With the loss function computed above, we derive the gradient of the loss function to be $\nabla \mathcal{L} := \partial \mathcal{L}(y, \hat{y})/\partial w$. At each iteration $t$, we compute $v_{t, \nabla \mathcal{L}} := \beta v_{t-1, \nabla \mathcal{L}} + (1-\beta) \nabla \mathcal{L}^2$ while $\beta$ is a tuning parameter. Note that the square term on $\nabla \mathcal{L}$ is element-wise multiplication. Then we can update the weights using $w_t := w_{t-1} - \eta \cdot \nabla \mathcal{L} / \sqrt{v_{t, \nabla \mathcal{L}}}$ while $\eta$ is learning rate. The value of learning rate $\eta$ is a tuning parameter and it is usually a very small number. This process starts with the loss function and goes back to the beginning to update the weights $w = \{w_1, w_2, w_3\}$. Hence, it earns the name backward propagation.


\subsection{Model Parameters} \label{model-parameters}
This section investigates the tuning parameter of proposed method. In the design of Interaction-based Convolutional Neural Network (ICNN), the window size and the level of stride are hyper-parameters. 

\textbf{Window Size.} Window size is the size of the local area that we narrow down to run the Backward Dropping Algorithm. For example, in the first Artificial Example in \S2.3, the data is sized $6 \times 6$. A window size of $2 \times 2$ means that we are starting with a local area that investigates the first row and the first two columns and the second row and the first two columns. For each row $i$ and each column $j$ in this $6 \times 6$ grid structure, a window size of $2 \times 2$ means a local area of the following 2 by 2 matrix 
\begin{equation*}
    \begin{bmatrix}
    (i, j) & (i, j+1) \\
    (i+1, j) & (i+1, j+1) \\
    \end{bmatrix}
\end{equation*}
For a 2 by 2 matrix, the Backward Dropping Algorithm will iteratively drop a variable amongst these 4 variables to compute I-score. The result will be a subset of this four variables.

We can also change the size of this window to be $3 \times 3$ which means we investigate the following matrix
\begin{equation*}
    \begin{bmatrix}
    (i, j) & (i, j+1) & (i, j+2) \\
    (i+1, j) & (i+1, j+1) & (i+1, j+2) \\
    (i+2, j) & (i+2, j+1) & (i+2, j+2) \\
    \end{bmatrix}
\end{equation*}
This means that the Backward Dropping Algorithm will start with 9 variables. After omitting the noisy variables within this set, the resulting predictive set will be a subset of these 9 variables.

\textbf{Stride Level.} The level of stride is how many rows or columns that get skipped over. This tuning parameter allows the algorithm to move faster but it makes sacrifice by skipping some variables. For example, we investigate stride level of 1 starting from row $i$ and column $j$. Assume we use a $2 \times 2$ window and let us start from $(i,j)$. We can visualize this action by using the following diagram
\begin{equation*}
\begin{array}{lll}
    \text{Original matrix: } &
    \begin{bmatrix}
    (i, j) & (i, j+1) \\
    (i+1, j) & (i+1, j+1) \\
    \end{bmatrix} \\
    \\
    \stackrel{\text{stride}=1}{\longrightarrow} & 
    \begin{bmatrix}
    (i, j+1) & (i, j+2) \\
    (i+1, j+1) & (i+1, j+2) \\
    \end{bmatrix} \\
\end{array}
\end{equation*}
If we are at the end of the column for a row, we move down by moving to the first column of the next row. For example, in a grid structure of size $6 \times 6$, assume we are in the last position in a certain row $i$. The action of stride level 1 can be taken using the following diagram
\begin{equation*}
\begin{array}{lll}
    \text{Original matrix } \\
    \text{in the end of a row: } \\
    \begin{bmatrix}
    (i, 5) & (i, 6) \\
    (i+1, 5) & (i+1, 6) \\
    \end{bmatrix}
    \stackrel{\text{stride}=1}{\longrightarrow}
    \begin{bmatrix}
    (i+1, 1) & (i+1, 2) \\
    (i+2, 1) & (i+2, 2) \\
    \end{bmatrix} \\
\end{array}
\end{equation*}

Again assume we are at row $i$ and column $j$. Suppose we set stride level to be 2 and we want to move down. This means we increase increment of 2 on the number of rows and the action is the following
\begin{equation*}
\begin{array}{lll}
    \text{Original matrix: } \\
    \begin{bmatrix}
    (i, j) & (i, j+1) \\
    (i+1, j) & (i+1, j+1) \\
    \end{bmatrix} &
    \stackrel{\text{stride}=2}{\longrightarrow} & 
    \begin{bmatrix}
    (i+2, j) & (i+2, j) \\
    (i+3, j) & (i+3, j) \\
    \end{bmatrix} \\
\end{array}
\end{equation*}
If this window happens to be in the final position of a row, then we move down by skipping one row and we start with the first column. If we have a grid structure of size $6 \times 6$, this action is shown in the following diagram
\begin{equation*}
\begin{array}{lll}
    \text{Original matrix } \\
    \text{in the end of a row: } \\
    \begin{bmatrix}
    (i, 5) & (i, 6) \\
    (i+1, 5) & (i+1, 6) \\
    \end{bmatrix}
    \stackrel{\text{stride}=2}{\longrightarrow}
    \begin{bmatrix}
    (i+2, 1) & (i+2, 2) \\
    (i+3, 1) & (i+3, 2) \\
    \end{bmatrix} \\
\end{array}
\end{equation*}

\textbf{Starting Point.} Another tuning parameter that we recommend to adjust is the starting point. The starting point represents the location of the first pixel that we start the proposed operation. The vanilla starting point is to start the rolling window from the pixel located at the first row and the first column. This can be illustrated in the following matrix
\begin{equation*}
\text{Starting point} = 1:
\begin{bmatrix}
    \color{red}{X_1} & \dots \\
    \vdots & \ddots \\
\end{bmatrix} \\
\end{equation*}
while the starting point is colored in red.

Alternatively, we can initiate the starting point to be at a higher level such as two or three. This allows algorithms to run more efficiently in large-scale data sets. For a simple example, in a matrix that is sized $6 \times 6$ (see \S2.3 for the first artificial example), we have the first row of variables to be $\{X_1, X_2, ..., X_6\}$ and the second row of variables to be $\{X_7, X_8, ... X_{12}\}$. At a starting point of 2, we start with $X_8$ to pass over the rolling window, because this variable sits at the position with the second row and the second column. This can be illustrated in the following matrix
\begin{equation*}
\text{Starting point} = 2:
\begin{bmatrix}
    X_1 & X_2 & \dots \\
    X_7 & \color{red}{X_8} & \dots \\
    X_{13} & \vdots & \ddots \\
\end{bmatrix}_{6 \times 6} \\
\end{equation*}
with the starting point colored in red.

This tuning parameter is particularly useful when the edge of the images are noisy and non-informative. For example, in Section \S4.1, we notice from training images of COVID-19 Chest X-ray data set that the margins of X-ray images are dark and do not have human body within the range of a few pixels length. This is an example when this tuning parameter can be helpful and it speeds up the training process.

\textbf{Computation of Dimensions.} The above discussion introduced the tuning parameters of window size, stride level, and starting point. These parameters update our input matrix and generate a new matrix with different sizes. Let us denote window size to be $w$, stride level to be $l$, and starting point to be $p$. Given an input matrix with size $s_\text{in}$ by $s_\text{in}$, the output matrix has new dimensions computed as the following
\begin{equation}\label{eq:new-dim}
    \floor{\frac{s_\text{in} - p - w + 1}{l} + 1} \times \floor{\frac{s_\text{in} - p - w + 1}{l} + 1}
\end{equation}
For simplicity of this investigation, we assume input matrices to be a square. In other words, in the application of this paper, we process the input images to have the same width and height, i.e. 128 by 128 pixels. For future work, this can be generalized into different shapes.

\subsection{Evaluation Metrics} \label{model-metrics}
This paper focuses on using Area-Under-Curve (or AUC) as the main evaluation metric. The AUC value is obtained from the Receiver Operating Characteristic (ROC) curve, a path constructed using different pairs of specificity and sensitivity. 

\textbf{Sensitivity and Specificity.} The notion of sensitivity is interchangeable with recall or true positive rate. In a simple two-class classification problem, the goal is to investigate covariate matrix $X$ in order to produce an estimated value of $Y$. From the output of a Neural Network model, the predicted values are always between 0 and 1, which acts as a probabilistic statement to describe the chance an observation is class 1 or 0. Given a threshold between 0 and 1, we can compute sensitivity to be the following
\begin{equation}\label{eq:definition-sensitivity}
    \begin{array}{rcl}
        \text{Sensitivity} 
        &=& \mathlarger{\frac{\text{True Positive}}{\text{Positive}}} \\
        &=& \mathlarger{\frac{\text{\# of Images Classified COVID-19 Correctly}}{\text{\# of COVID-19 Images}}}
    \end{array}
\end{equation}
On the other hand, specificity is also used to create ROC curve. Given a certain threshold between 0 and 1, we can compute specificity using the following
\begin{equation}\label{eq:definition-specificity}
    \begin{array}{rcl}
        \text{Specificity}
        &=& \mathlarger{\frac{\text{True Negative}}{\text{Negative}}} \\
        &=& \mathlarger{\frac{\text{\# of Images Classified Non-COVID Correctly}}{\text{\# of Non-COVID Images}}}
    \end{array}
\end{equation}
Given different thresholds, a list of pair of sensitivity and specificity can be created. The Area-Under-Curve (AUC) is the area under the path plotted using pairs of sensitivity and specificity that is generated using different thresholds.
\begin{figure}[H]
    \centering
    \caption{This figure presents two paths of ROC curves. We call them ROC1 and ROC2. For each ROC curve, we can compute AUC value. From ROC1, we can compute AUC1. From ROC2, we can compute AUC2. The mistake discussed in the paragraph is reflected by a reduction of AUC values from path 1 to path 2.}
    \begin{tabular}{cc}
        \includegraphics[width=0.6\textwidth]{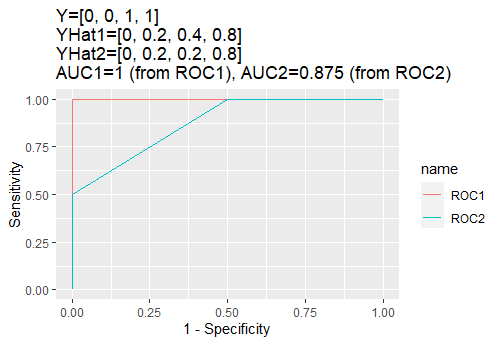} \\
    \end{tabular}
    \label{fig:auc-explained}
\end{figure}

\textbf{Area-Under-Curve (AUC).} The AUC value is a single number derived from a predicted probability by a classification rule and the true label \cite{hand2009measuring}. Given a vector of true label $Y$ and a vector of predicted probability $\hat{Y}$, we can use statistical package ``pROC''\footnote{The package is called ``Display and Analyze ROC Curves''. Source: \url{https://github.com/xrobin/pROC}} to assist this computation. The package uses automatically generated thresholds to convert $\hat{Y}$ into binary format. For example, we can use threshold of $t_1=0.3$ to convert a vector of predicted probabilities $\hat{Y} = [0, 0.2, 0.4, 0.8]$ into binary form by writing $\hat{Y}_{t_1} = \mathbb{1}(\hat{Y} > t_1) = [0, 0, 1, 1]$. Let us assume the true label to be $Y = [0, 0, 1, 1]$. Thus, we can compute specificity to be 1 and sensitivity to be 1. We can then change threshold to a different value to compute another pair of specificity and sensitivity. By tracking all pairs of specificity and sensitivity, we can generate a curve called Receiver Operating Characteristic (ROC) \cite{hand2009measuring}. The value of Area-Under-Curve is exactly the area under the ROC. Assume the predicted probability to have some mistakes. In other words, let us assume the predicted probability vector to be $\hat{Y} = [0, 0.2, 0.2, 0.8]$. It is not possible for two observations to have the same probability predicted while they come from different classes. Therefore, there must be a mistake in one of them. This information is reflected using the same procedure. Please see Table \ref{fig:auc-explained}.

\subsection{Simulation with Artificial Examples}
In this section, we illustrate proposed methodologies on the following artificial examples.

\subsubsection{Artificial Example: Variable/Feature Investigation}
The first artificial example we demonstrate the procedure to construct Interaction-based Convolutional Layer and engineer Interaction-based Features. 

Let us consider the following experiment. We create an artificial data set with 500 data points in training set and 10,000 data points in testing set. These observations are randomly drawn from $\text{Bernoulli}(1/2)$ independently. In other words, we independently generate $X_i \sim \text{Bernoulli}(1/2)$ while $i = \{1, 2, 3, ..., 36\}$. We define the underlying model to be the following (here known as model \ref{eq:artificial_example_1}),
\begin{equation}\label{eq:artificial_example_1}
Y = \left\{
\begin{array}{ll}
X_1 + X_2 & (\text{mod } 2) \text{ with prob. } 0.5 \\
X_3 + X_4 + X_5 & (\text{mod } 2) \text{ with prob. } 0.5 \\
\end{array}
\right.
\end{equation}
This model \ref{eq:artificial_example_1} is a two-module example. The first module is a two-way interaction in modulo 2. The second module is a three-way interaction in modulo 2. The response variable $Y$ is defined to be the first module 50\% and the second module 50\%. In this setup, the individual explanatory variable does not have any predictive power on the response variable $Y$.

\textbf{Scenario I.} Assume the statistician knows the model in this simulation. This means he is fully aware of $S_1 = \{X_1, X_2\}$ to be an important variable set and $S_2 = \{X_3, X_4, X_5\}$ to be the other. In other words, he can use the first module as a predictor to make predictions on response variable $Y$. He is able to compute the theoretical prediction rate of the first variable set to be 75\%. This is because the response variable is defined to be the first variable module $S_1 = \{X_1, X_2\}$ exactly 50\% of the times so $S_1$ is able to guess $Y$ correctly at least 50\% of the times. The rest 50\% of the times the response variable is defined as the second variable module $S_2 = \{X_3, X_4, X_5\}$. Since there is no marginal signal, the performance is exactly like random guessing and hence only get the rest of the occurrences correct 50\% of the times. In other words, assuming training sample size has $n$ data points, we can write the following
\begin{equation}\label{eq:artificial_example_1_one_module_bayesrate}
\begin{array}{rcl}
\theta_c(S_1) 
&=& \theta_c(\{X_1,X_2\}) \\
&=& \frac{1}{n} \sum_i^n \mathds{1}(\hat{Y} = Y) \\
&=& \frac{1}{n} \sum_i^n \mathds{1}(\underbrace{(X_1 + X_2)}_\text{(mod 2)} = Y) \\
&=& 50\% + 50\% \cdot 50\% \\
&=& 75\% \\
\end{array}
\end{equation}

This result can be extended to the other variable module as well. If this statistician uses the second variable module $S_2 = \{X_3, X_4, X_5\}$, then a similar calculation can be carried out in the following.
\begin{equation}\label{eq:artificial_example_1_one_module_bayesrate_s2}
\begin{array}{rcl}
\theta_c(S_2) 
&=& \theta_c(\{X_3,X_4,X_5\}) \\
&=& \frac{1}{n} \sum_i^n \mathds{1}(\hat{Y} = Y) \\
&=& \frac{1}{n} \sum_i^n \mathds{1}(\underbrace{(X_3 + X_4 + X_5)}_\text{(mod 2)} = Y) \\
&=& 50\% + 50\% \cdot 50\% \\
&=& 75\% \\
\end{array}
\end{equation}

The theoretical prediction rate for model \ref{eq:artificial_example_1} is the maximum percentage accuracy delivered by $S_1$ and $S_2$. In this case, we have $\max(\theta_c(S_1), \theta_c(S_2)) = 0.75$.

\begin{figure}
    \centering
    \caption{This figure demonstrate the network architecture in simulated data. The artificial data has $6 \times 6 = 36$ variables which can be arranged in a grid structure with shape 6 by 6. We use a window size that is 2 by 2. By passing this window from top left corner of the original 6-by-6 matrix, we would create a new matrix that has dimension 5 by 5. The number 32 is the number of units in the hidden layer. In this example, there is one hidden layer which is sufficient for the dimension of the data in the example.}
    \label{fig:iCNN-simulated}
    \includegraphics[width=0.25\textwidth]{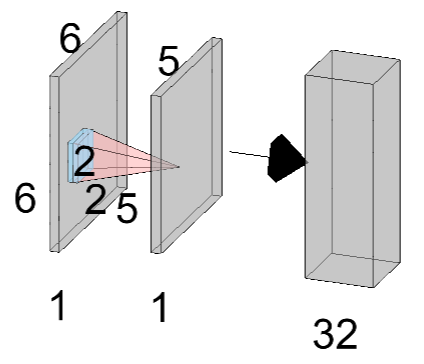}
\end{figure}

\textbf{Scenario II.} In practice, it is often the case that we do not observe exactly what the underlying model is in a given data set. The recommendation is to use I-score to select the important local information and then create an interaction-based feature based on the selected variable. The diagram for this action can be seen in Figure \ref{fig:iCNN-simulated}.

Classical procedure of Convolutional Neural Network starts with many pre-defined filters (or kernels) from a previous data set. The size of this filter can be $2 \times 2$, $3 \times 3$, or any other higher dimension. The procedure starts with the first row and the first column. Then the process takes the filter to pass it over across rows and columns in the image. The proposed methodology shares similar characteristics. However, instead of using a pre-trained filter, we apply the Backward Dropping Algorithm in this local area. 

\textbf{Constructing the Proposed Convolutional Layer.} In the proposed artificial example, we have a data set with 36 variables. This means each observation can be reshaped into $6 \times 6$ grid structure. In other words, each observation may be considered as an image. The first row and the first column is variable $X_1$. The first row and the last column will be variable $X_6$. This means we can arrange these variables into the following structure
\begin{equation*}
    \begin{array}{ll}
    \text{Originally: } & \\
    \{X_1, X_2, ..., X_{36}\}
    \longrightarrow & 
    \underbrace{\begin{bmatrix}
    \color{red}{X_1} & \color{red}{X_2} & X_3 & \dots & X_6 \\
    \color{red}{X_7} & \color{red}{X_8} & X_9 & \dots & X_{12} \\
    \vdots & \vdots & \ddots & \\
    X_{31} & X_{32} & X_{33} & \dots & X_{36} \\
    \end{bmatrix}}_\text{6 by 6} \\
\end{array}
\end{equation*}
and the red colored 4 variables are first used to the Backward Dropping Algorithm. If the window size is $2 \times 2$, we would use $\{X_1, X_2, X_7, X_8\}$ and execute the Backward Dropping Algorithm on the variables within this window.

Notice that in the diagram (see Figure \ref{fig:iCNN-simulated}) the blue region indicates a local area where we run the Backward Dropping Algorithm. The pink region indicates the engineering workflow of building Interaction-based Feature using equation \ref{eq:interaction-based-features-general-form}. Each window we run Backward Dropping Algorithm. The procedure adopts the steps introduced in \S2.2. The steps can be seen in Table \ref{tab:iCNN-toy-example-1-bda-illustration}. First, we start with all 4 variables which are $\{X_1, X_2, X_7, X_8\}$. For each step, we take turns dropping one variable and compute the I-score for the remaining variables. It can be seen that $X_8$ should be dropped because I-score raises from 160.18 in step 1 to 319.65 to step 2. We conduct the same procedure in step 2 and realize that $X_7$ needs to be dropped. Then I-score can increase to 638.17. We can keep dropping variables, but statistician realizes that I-score is the highest when there is only two variables, $\{X_1, X_2\}$, left. Hence, we put an up-arrow at the step where it indicates the most optimal variable selection (see up-arrow in Table \ref{tab:iCNN-toy-example-1-bda-illustration}). In this experiment, the most optimal selection is the variable module $\{X_1, X_2\}$ with an I-score of 638.17.

\begin{table}[H]
    \centering
    \caption{The table illustrates the steps of the Backward Dropping Algorithm in a $2 \times 2$ window.}
    \begin{tabular}{lllll}
        \toprule
        Step    & 1      & 2      & 3      & 4 \\
        Drop    & Start  & $X_8$  & $X_7$  & $X_2$ \\
        \hline
        I-score & 160.18 & 319.65 & 638.17 & 0.65 \\
                & $X_1$  & $X_1$  & $X_1$  & $X_1$ \\
                & $X_2$  & $X_2$  & $X_2$  & \\
                & $X_7$  & $X_7$ \\
                & $X_8$  &  \\
        \hline
        Best result:  & $\{X_1,X_2\}$ \\
        \bottomrule
    \end{tabular}
    \label{tab:iCNN-toy-example-1-bda-illustration}
\end{table}

Each local area with a window size of $2 \times 2$ we discuss in the above how to finely select the important variable. Similar to the flavor of standard procedure in Convolutional Neural Network to use filter on a local area of an image, the proposed method above also investigate local information. However, instead of relying on many filters that are pre-defined, the proposed approach uses I-score, a model free influence measure that directly associates predictivity. After the variable module is selected within a local window, we adopts formula \ref{eq:interaction-based-features-general-form} to create Interaction-based Features. For example, in the local window above, we narrow down to variable module $\{X_1, X_2\}$. This means we can create the first Interaction-based Feature to be the following
\begin{equation*}
    X^\dagger_1 := 
    \left\{
    \begin{array}{lcl}
        \bar{y}_1 & \text{if obs. falls in } X_1 = 1, X_2 = 1 \\
        \bar{y}_2 & \text{if obs. falls in } X_1 = 1, X_2 = 0 \\
        \bar{y}_3 & \text{if obs. falls in } X_1 = 0, X_2 = 1 \\
        \bar{y}_4 & \text{if obs. falls in } X_1 = 0, X_2 = 0 \\
    \end{array}
    \right.
\end{equation*}
while $j = \{1, 2, 3, 4\}$ and this corresponds to the $2^2 = 4$ partitions generated by $\{X_1, X_2\}$. This gives us the first predictor to allow us to build classifier and to make predictions. 

The experimental results from the first artificial example can be seen in Table \ref{tab:artificial-example-1-module-information-2x2}, Table \ref{tab:artificial-example-1-module-information-3x3} and Table \ref{tab:iCNN-toy-example-1-results}. Suppose a statistician knows the model. The theoretical prediction rate is 75\% (which is derived using formula \ref{eq:artificial_example_1_one_module_bayesrate} and \ref{eq:artificial_example_1_one_module_bayesrate_s2}). In reality, we suppose that a statistician have no knowledge of the data set. In this case, he can directly proceed using Neural Network or even Convolutional Neural Network to make predictions. However, without the correct model specification the performance for these models are sub-par. We can see that Neural Network and Convolutional Neural Network both performed 50\% on test set. This performance is rather like random guessing. Alternatively, this statistician can proceed the experiment using the proposed methods. First, one can generate Interaction-based Convolutional Layer. Instead of using many pre-trained filters for feature mapping, the proposed method adopts I-score and the Backward Dropping Algorithm to construct $X^\dagger_{\{X_1, X_2\}}$, $X^\dagger_{\{X_8, X_9, X_{21}\}}$, and so on. The proposed methodology also has an exact I-score that is associated to each variable module which allow us to rank feature importance. With I-score, it is obvious that the first variable module $X^\dagger_{\{X_1,X_2\}}$ is much more influential than the other variable modules. 

\begin{table}
    \centering
    \caption{This figure presents Interaction-based Conv. Layer. For artificial data set with $6^2 = 36$ variables and a window size of $2 \times 2$, we pass the window from top left corner down to the bottom right corner. Each particular location, we have 4 variables to run BDA. Before each observation has 36 features that can be sized $6 \times 6$. Afterwards, each observation has 25 new features that has the shape of $5 \times 5$. In other words, we create $X^\dagger_b$ while $b = \{1, 2, ..., 25\}$. The ``*'' symbol indicates extremely influential variable module(s). For each variable module, we also present the AUC value (see \S4.4 for detailed discussion of AUC values) assuming a classifier is built using this variable module alone.}
    \resizebox{\textwidth}{!}{
    \begin{tabular}{cccc|cccc}
    \toprule
        New Mod. & Variables & I-score & AUC & New Mod. & Variables & I-score & AUC \\
        \hline
        $X^\dagger_{1}$*	&	X1, X2			&	638.17	& 0.75 &	$X^\dagger_{14}$	&	X28			&	1.3729 & 0.50	\\
        $X^\dagger_{2}$	&	X7				&	1.2162	& 0.50 &	$X^\dagger_{15}$	&	X28			&	1.3729 & 0.50	\\
        $X^\dagger_{3}$	&	X13, X20			&	2.3597	& 0.51 &	$X^\dagger_{16}$	&	X11			&	0.2347 & 0.50	\\
        $X^\dagger_{4}$	&	X19, X20, X26		&	0.7218	& 0.50 &	$X^\dagger_{17}$	&	X11			&	0.2347 & 0.50	\\
        $X^\dagger_{5}$	&	X26, X31			&	2.6751	& 0.50 &	$X^\dagger_{18}$	&	X16, X22		&	0.0777 & 0.51	\\
        $X^\dagger_{6}$	&	X8, X9			&	0.5067	& 0.49 &	$X^\dagger_{19}$	&	X28			&	1.3729 & 0.51	\\
        $X^\dagger_{7}$	&	X8, X9			&	0.5067	& 0.50 &	$X^\dagger_{20}$	&	X28			&	1.3729 & 0.51	\\
        $X^\dagger_{8}$	&	X15, X21			&	1.8013	& 0.50 &	$X^\dagger_{21}$	&	X6, X12		&	0.4378 & 0.49	\\
        $X^\dagger_{9}$	&	X20, X21, X26, X27	&	0.7554	& 0.50 &	$X^\dagger_{22}$	&	X11, X12		&	0.6184 & 0.51	\\
        $X^\dagger_{10}$	&	X27, X32			&	1.017 & 0.50	&	$X^\dagger_{23}$	&	X18, X24		&	1.3814 & 0.51	\\
        $X^\dagger_{11}$	&	X9, X10			&	0.6894	& 0.50 &	$X^\dagger_{24}$	&	X23, X24, X29	&	0.8788 & 0.51	\\
        $X^\dagger_{12}$	&	X9, X10, X15		&	0.9346 & 0.51	&	$X^\dagger_{25}$	&	X30, X35		&	1.2105 & 0.51	\\
        $X^\dagger_{13}$	&	X15, X16, X21, X22	&	1.0933	& 0.50 &		&				&		\\
    \bottomrule
    \end{tabular} }
    \label{tab:artificial-example-1-module-information-2x2}
\end{table}

We pursue using window size $2 \times 2$. In a data set with $6^2 = 36$ variables, this gives us $(6-2+1)^2 = 25$ variable modules. Using these 25 variable modules we are able to build a classifier and delivers the performance of 76\% AUC (see \S4.4 for detailed discussion of AUC values) on test set, much higher than 50\% from before. We can also use a window size that is $3 \times 3$. This gives us $(6-3+1)^2 = 16$ variable modules. These 16 modules allow us to build another classifier using Neural Network and delivers 76\% AUC on test set as well.

The detailed variable modules and their corresponding I-score are exhibited in Table \ref{tab:artificial-example-1-module-information-2x2}. In this table, we present three columns: New Modules, Variables, associated I-score, and corresponding AUC. The new module is constructed using \ref{eq:interaction-based-features-general-form}. The variables come from BDA. The I-score of a variable set is computed using equation \ref{eq:iscore}. For computation of AUC, please see section \S4.4. In the artificial data set that has $6 \times 6 = 36$ variables, a window size of $2 \times 2$ allows us to generate $(6-2+1)^2 = 25$ variable modules. Each module is created using selected variables from a local $2 \times 2$ window after using the Backward Dropping Algorithm and the module is generated using formula \ref{eq:interaction-based-features-general-form}. We easily observe that the variable module $X^\dagger_1$ that is formed based on $\{X_1, X_2\}$ is the most influential candidate because it directly links to predictivity. This can be confirmed with AUC value if this module is used to build a classifier by itself.

A different size of window can be selected. In this artificial example, we attempted the size $3 \times 3$. For a data with $6 \times 6 = 36$ variables, a window size of $3 \times 3$ allows us to create $(6-3+1)^2 = 16$ variable modules. The results of this experiment can be seen in Table \ref{tab:artificial-example-1-module-information-3x3}. In Table \ref{tab:artificial-example-1-module-information-3x3}, we observe a slightly different sets of variable modules selected. We notice that the most influential variable modules $X^\dagger_1$ that is based on $\{X_1, X_2\}$ remains the same. In other words, if a local section of the data has real predictive power, the proposed method will almost always pick out the information. 

\begin{table}
    \centering
    \caption{This figure presents Interaction-based Conv. Layer. For artificial data set with $6^2 = 36$ variables and a window size of $3 \times 3$, we create 16 new features that can be shaped into size $4 \times 4$. The procedure of generating these 16 new features take the same procedure as Table \ref{tab:artificial-example-1-module-information-2x2}. The only thing different is the window size, i.e. here we use $3 \times 3$. In other words, we are able to create $X^\dagger_b$ while $b = \{1, 2, ..., 16\}$. The ``*'' symbol indicates extremely influential variable module(s). For each variable module, we also present the AUC value (see \S4.4 for detailed discussion of AUC values) assuming a classifier is built using this variable module alone.}
    \resizebox{\textwidth}{!}{%
    \begin{tabular}{cccc | cccc}
    \toprule
        New Mod. & Variables & I-score & AUC & New Mod. & Variables & I-score & AUC \\
        \hline 
        $X^\dagger_{1}$* & X1, X2                  & 638.17 & 0.75 & $X^\dagger_{9}$* & X3, X4, X5              & 350.2429 & 0.75\\
        $X^\dagger_{2}$ & X8, X9, X13, X15, X21   & 0.6344 & 0.50 & $X^\dagger_{10}$ & X11                    & 0.2347 & 0.51 \\
        $X^\dagger_{3}$ & X19, X21, X25           & 1.4386 & 0.50 & $X^\dagger_{11}$ & X16, X17, X21, X22     & 0.8097 & 0.51 \\
        $X^\dagger_{4}$ & X19, X21, X25           & 1.4386 & 0.50 & $X^\dagger_{12}$ & X28                    & 1.3729 & 0.51 \\
        $X^\dagger_{5}$ & X8, X9                  & 0.5067 & 0.51 & $X^\dagger_{13}$ & X11                    & 0.2347 & 0.51 \\
        $X^\dagger_{6}$ & X10, X15, X21           & 0.9883 & 0.50 & $X^\dagger_{14}$ & X18, X24               & 1.3814 & 0.51 \\
        $X^\dagger_{7}$ & X14, X15, X21, X22, X26 & 0.9816 & 0.50 & $X^\dagger_{15}$ & X18, X24               & 1.3814 & 0.51 \\
        $X^\dagger_{8}$ & X20, X32, X33           & 2.0205 & 0.50 & $X^\dagger_{16}$ & X22, X23, X24, X29, X34 & 1.5684 & 0.51 \\
    \bottomrule
    \end{tabular} }
    \label{tab:artificial-example-1-module-information-3x3}
\end{table}

This can be verified using AUC value (see \S4.4 for detailed discussion of AUC values) associated with the variable module. Since we pass over a window that is sized $3 \times 3$, we are able to sometimes screen for higher order interaction. As discussed before, we have I-score presented in Table \ref{tab:artificial-example-1-module-information-3x3} so that feature ranking is possible if desired. From this table, we observe that another module $X^\dagger_9$ that is constructed using $\{X_3, X_4, X_5\}$ is also very important. Though it has a lower I-score than the first module $X^\dagger_1$, it has I-score higher than any other candidates in the data set. 

\textbf{Prediction.} After the variable modules (the variables named $X^\dagger$'s) are constructed, we can proceed with building neural network classifier. In Table \ref{tab:artificial-example-1-module-information-2x2}, there are 25 variable modules, i.e. $\{X^\dagger_1, ..., X^\dagger_{25}\}$. We use these variable modules as the input layer of a neural network architecture. In other words, the input layer has 25 neurons (these neurons are exactly the 25 variable modules named $X^\dagger$'s). For variable modules in Table \ref{tab:artificial-example-1-module-information-3x3}, the input layer consists of 16 neurons, which are $\{X^\dagger_1, ..., X^\dagger_{16}\}$. Since we are dealing with a rather small size of input layer (only 25), there is no hidden layer recommended. The output layer can simply be one neuron and the final outcome $\hat{Y}$ can be computed using a sigmoid function as the following
\begin{equation}\label{eq:artificial-example1-2x2-yhat}
    \hat{Y} := 1/(1+\exp(- \sum_j^{25} (w_jX_j^\dagger))
\end{equation}
assuming we use the 25 variable modules (the $X^\dagger$'s) in Table \ref{tab:artificial-example-1-module-information-2x2}. Alternatively, suppose we use the 16 variables modules, $\{X^\dagger_1, ..., X^\dagger_{16}\}$, in Table \ref{tab:artificial-example-1-module-information-3x3}. We can compute $\hat{Y}$ using the following
\begin{equation}\label{eq:artificial-example1-3x3-yhat}
    \hat{Y} := 1/(1+\exp(- \sum_j^{16} (w_jX_j^\dagger))
\end{equation}
assuming the same activation function to be sigmoid (same as \ref{eq:artificial-example1-2x2-yhat}). This architecture is called a forward propagation (a standard architecture widely used in neural network). This is discussed in \S4.2 and please read the architecture \ref{fig:nn-architecture} for detailed description. We assume there is no bias term. Then the objective is to minimize the loss between $Y$ and $\hat{Y}$. Hence, we may write $\min_{\vec{w}} \mathcal{L}(Y, \hat{Y})$. We can simply choose the loss function to be ``square-error'', which means $\mathcal{L}(Y, \hat{Y}) = \sum_{i=1}^n (Y_i - \hat{Y}_i)^2$. In order to minimize this loss function, an optimizer algorithm is required and we can choose the standard gradient descent algorithm.

The results for the first artificial example is presented in Table \ref{tab:iCNN-toy-example-1-results}. We discussed in Scenario I that the theoretical prediction rate is 75\% if a statistician knows the model. In practice, we assume that the statistician has no knowledge of the model. In this case, the standard approach is to directly work with all variables. This means one can build a classifier such as using Neural Network. In this example, we observe the data set has number of variables to be $6 \times 6 = 36$. This means one can also adopt Convolutional Neural Network by reshaping the input dimension for an observation from $1 \times 36$ into $6 \times 6$. Both Neural Network and Convolutional Neural Network deliver poor performance, i.e. approximately 50\% AUC. This is because the underlying model is in binary form and there is no marginal signal. In addition, the in-sample training set has only 500 observations which further raise difficulty for complex architecture such as deep learning models. However, this issue can be resolved if we have the correct model specification. Based on the proposed methodology, we are able to create the variable modules presented in Table \ref{tab:artificial-example-1-module-information-2x2} and Table \ref{tab:artificial-example-1-module-information-3x3}. Their performance of test set is presented in Table \ref{tab:iCNN-toy-example-1-results}. A window size of $2 \times 2$ produces 25 variable modules and deliver AUC of 75\%. A window size of $3 \times 3$ produces 16 variable modules and delivers AUC of 76\%. These results approximate to 75\% theoretical prediction rate.

\begin{table}
    \centering
    \caption{The table presents performance of the this simulation. In this simulation, we define the underlying model to be \ref{eq:artificial_example_1}. The theoretical prediction rate is 75\%. The conventional model such as Net-3 and LeNet-5 do not perform well in this example \cite{yann1989} \cite{yann1998mnist}. The proposed method that uses I-score has prediction performance that is close to the theoretical prediction rate (see \S4.4 for detailed discussion of AUC values). The notation ``NN'' stands for neural network. This is discussed in \S4.2 and please read the architecture \ref{fig:nn-architecture} for detailed description.}
    \begin{tabular}{lcc}
        \toprule
        Algorithm              & Test AUC \\
        \hline \hline
        Theoretical Prediction & 0.75 \\
        \hline
        Net-3    & 0.50 \\
        LeNet-5  & 0.50 \\
        \hline
        Interaction-based Conv. Layer: \\
        window size: $2 \times 2$ \\
        (25 modules listed in Table \ref{tab:artificial-example-1-module-information-2x2}) \\
        Using $\hat{Y}$ defined in equation \ref{eq:artificial-example1-2x2-yhat} \\
        Interaction-based Conv. Layer + NN & 0.75 \\
        \hline
        Interaction-based Conv. Layer: \\
        window size: $3 \times 3$ \\
        (16 modules listed in Table \ref{tab:artificial-example-1-module-information-3x3}) \\
        Using $\hat{Y}$ defined in equation \ref{eq:artificial-example1-3x3-yhat} \\
        Interaction-based Conv. Layer + NN & 0.76 \\
        \bottomrule
    \end{tabular}
    \label{tab:iCNN-toy-example-1-results}
\end{table}

\subsection{Data}\label{data-source}

The source of the data is from the work by \cite{Minaee2020}. The link of their work can be accessed here: \url{https://pubmed.ncbi.nlm.nih.gov/32781377/}. The authors \cite{Minaee2020} made the dataset publicly available for research community at \url{https://github.com/shervinmin/DeepCovid.git}. We download 576 COVID images directly from their work. We also collect 2,000 healthy chest X-ray images from their database and use these images as non-COVID case. The work by \cite{Minaee2020} also has other diseases such as Pneumonia in the data. The goal for our work is to understand the difference between COVID chest X-ray images and healthy images. Therefore, we do not take other diseases into consideration. We present the summary of these information in Table \ref{tab:covid-data-size-summary}. We first randomly select 60 images each from COVID class and non-COVID class and use these images as out-of-sample test set. We do not touch the test set until the very end after training and validation is completed. For the remaining images, we use data augmentation technique by adding noises drawn from normal distribution. This gives us totalled of 5,000 COVID cases and 5,000 non-COVID cases. These 10,000 images consist of the training and validating sets (short for tr. and val. in the Table \ref{tab:covid-data-size-summary}). These 10,000 images have two classes: COVID and non-COVID. of COVID-19 diseases on CT scans have been proposed. Bai et al. (2020) provided the model output to radiologists, and demonstrated that AI-assistance significantly improved radiologist diagnostic accuracy from 85\% to 90\% in distinguishing COVID classes from non-COVID classes \cite{bai2020artificial}. Minaee et al. (2020) \cite{Minaee2020} have demonstrated using deep CNNs including ResNet18, ResNet50, DenseNet-121 to classify COVID-19 disease using X-ray images. They have achieved sensitivity rate of 98\% and specificity of 90\%. on 5,000 Chest X-ray images.

\subsection{Models}

We discuss updated versions of \textbf{Model 1} in the following. \label{updated-models}

\textbf{Model 2.} This model builds up the architecture of Model 1. The only difference is that there is one hidden layer with 64 units (or neurons). We fully connect each variable in 1st Conv. layer with each neurons in the hidden layer and afterwards we fully connect the hidden layer with the output layer. This means that from 1st Conv. Layer to the hidden layer there are $3,721 \times 64 = 238,144$ parameters. From the hidden layer of 64 neurons to output layer with 2 units, there are $64 \times 2 = 128$ parameters. This means in total there are $238,144 + 128 = 238,272$ parameters. The performance for this architecture is 99.7\%. The design of this one hidden layer with 64 units reduced the error rate from 1.5\% in Model 1 to 0.3\% in Model 2, which is 80\% error reduction.

\textbf{Model 3.} This model aims to build two Interaction-based Convolutional Layer. The 1st Conv. Layer uses the set of parameters in $\triangle$ and the 2nd Conv. Layer uses the set of parameters in $\square$. From the 1st Conv. Layer in Model 1, we are left with $61 \times 61 = 3,721$ variables. Using the parameters in $\square$, we have new matrix with size $\floor{(61-1-2+1)/2+1} \times \floor{(61-1-2+1)/2+1} = 30 \times 30 = 900$ variables. These 900 variables can be input layer and we can directly pass these 900 variables into the output layer for making predictions. In other words, the output layer has $900 \times 2 = 1,800$ parameters. The test set AUC value is 97.0\%.

\textbf{Model 4.} This model is the deepest amongst all six models. More specifically, Model 4 has two Interaction-based Convolutional Layers and one hidden layer. From the 2nd Conv. Layer in Model 3, we are left with 900 variables. The architecture has one hidden layer with 64 units. The 900 variables are fully connected with the hidden layer which create $900 \times 64 = 57,600$ variables. From the hidden layer with 64 units to output layer with 2 units, there are $64 \times 2 = 128$ parameters. In total, there are $57,600 + 128 = 57,728$ parameters. The prediction performance is 99.6\% on test set.

\textbf{Model 5.} Both Model 5 and Model 6 build wider convolutional layers instead of aiming for depth. Model 5 has a concatenated of features from both convolutional layers. This means the architecture takes the 7,442 variables from the 1st Conv. Layer and 900 variables from the 2nd Conv. Layer from previous models together as one large convolutional layer. In other words, Model 5 has 1st Conv. Layer of $3,721 + 900 = 4,621$ variables. These 4,621 variables can be used directly to be fed into the output layer with 2 units. In total, this architecture creates $4,621 \times 2 = 9,242$ parameters with test set performance to be 98.3\%.

\textbf{Model 6.} The last model, Model 6, is just as wide as the previous model, Model 5. Model 6 also has a first convolutional layer that is concatenation of features. It has $3,721 + 900 = 4,621$ variables. In addition to Model 5, it has one hidden layer with 64 units. We fully connect the convolutional layer of 4,621 variables with the hidden layer of 64 variables. This gives us $4,621 \times 64 = 295,744$ parameters. The hidden layer with 64 units are then fully connected with the output layer which gives us $64 \times 2 = 128$ parameters. In total, the model has $295,744 + 128 = 295,872$ parameters. This model has the highest AUC value on test set, i.e. 99.8\%.

\subsection{Visualization}

Discussion for Figure \ref{fig:main-diagram}.\label{fig:main-diagram-appendix}

This executive diagram summarizes the key components of the proposed methods of this paper. We start with the COVID-19 Image Data. With a small rolling window defined, we execute the Backward Dropping Algorithm to select the important features within this window. For example, the rolling window may cover 4 variables, $\{X_1, X_2, X_3, X_4\}$ and BDA could select $\{X_1, X_2\}$ as a variable module. Then we can construct a new variable using the technique of Interaction-based Feature Engineer (see the construction of $X^\dagger$ in equation \ref{eq:interaction-based-features-general-form} to appreciate this new design). In other words, using the selected variables $X_1$ and $X_2$, we construct $X^\dagger$. The procedure of BDA is illustrated in the bottom left corner of the figure (we use a 2 by 2 window for simplicity). We set the starting point to be 12 (i.e. start from the pixel at the 12th row and the 12th column). From data (size of 128 by 128) to the 1st layer (58 by 58), this gives us a new dimension that is computed by $\floor{(128-12-2+1)/2 + 1} = 58$ (see equation \ref{eq:new-dim}). We can repeat the process in the 2nd Layer and the 3rd Layer. After the 3rd Layer, we shrink the dimension to 14 by 14 (which gives us 196 new variable modules, i.e. the new $X^\dagger$'s). We fully connect these 196 variable modules with the 10 neurons in the hidden layer (in practice the number of hidden layer and the number neurons are tuning parameters). This novel design is fundamentally different than the conventional practice of using pre-trained filters, because it proposes to use an explainable measure I-score to extract and build information directly from images in the training data. For each local variable (in the data it is referring to pixels and in the layers it is referred as variables), we compute the I-score values and also the AUC (see \S4.4 for detailed discussion of AUC values) for that variable (using this variable as a predictor alone). We observe that the I-score value fully represents the predictivity of each local variable which can be confirmed by the variable's AUC value. The color spectrum of both I-score and AUC are presented in the bottom part of the diagram. We observe I-score values exhibiting paralleled behavior with AUC values. This means the variables with high I-score values have high AUC values which indicates strong predictive power for the information in that location. This design heavily rely on I-score and has an architecture that is interpretable at each location of the image at each convolutional layer. More importantly, the proposed design satisfies all three dimensions ($\mathcal{D}1$, $\mathcal{D}2$, and $\mathcal{D}3$ in \S1 Introduction) of the definition of interpretability and explainability. 

Discussion for Figure \ref{fig:covid-data-conv-layer-samplewise-plot}.\label{fig:covid-data-conv-layer-samplewise-plot-appendix}

This figure presents visualization summary for 10 randomly sampled images from COVID class and non-COVID class (each has 10). Panel A is for COVID patients and Panel B is non-COVID people. The first row plots the original images that are sized 128 by 128. The 1st Conv. Layer generates $61 \times 61 = 3,721$ new variables. We plot the \textbf{same} 10 images from both classes using these 3,721 variables in the second row. We also print the predicted COVID probabilities on top left corner of each image. The 2nd Conv. Layer generate $30 \times 30 = 900$ variables. We plot the same 10 image samples from both classes using these 900 variables in the third row. We also print the predicted COVID probabilities on top left corner of each image assuming using only these 900 variables as predictors. The plot of the original images for COVID-19 patients has grey and cloudy textures in chest area. This is due to inflammatory fluid when patients exhibit pneumonia-like symptoms. This shaded (as seen in Panel A) prevents us from observing the clear location of lungs. This is different in Panel B where the lung areas are dark and almost black which means the lung is filled with air (i.e. normal cases). The black white contrast in the two panels is directly related to how much inflammatory fluid there is in human lungs which translate to greyscale on pictures. The same contrast can be seen using the new variables (these are $X^\dagger$'s based on equation \ref{eq:interaction-based-features-general-form}) in the 1st Conv. Layer (sized 61 by 61). For COVID-19 patients, the lung area is cloudy and unclear while the healthy cases it is clearly visible.

Discussion for Figure \ref{fig:covid-data-conv-layer-samplewise-plot}.

\textbf{Original Images to 1st Conv. Layer.} The input images are sized 128 by 128. With the 1st Conv. Layer constructed, we have $61 \times 61 = 3,721$ new variables. We trace back to the same samples as shown in the first row of Figure \ref{fig:covid-data-conv-layer-samplewise-plot} and use these 3,721 variables only. When we plot these samples with these new variables, we resize them back in matrix form of 61 by 61. Panel A is for COVID class and Panel B is for non-COVID class. In addition, we use Model 1 in Table \ref{tab:covid-data-experimental-results} to produce the texts that states predicted probability of COVID class. The red color implies ground truth to be COVID class (Panel A) and the green color implies ground truth to be non-COVID class (Panel B). 

\textbf{1st Conv. Layer. to 2nd Conv. Layer.} From the resulting matrix of the 1st Conv. Layer, we are left with 3,721 variables. We go through the proposed design in Table \ref{tab:covid-data-experimental-results} and we create a new convolutional layer, i.e. 2nd Conv. Layer. This new layer has $30 \times 30 = 900$ variables. We take the same 10 sampled images from before and we use these 900 variables to present these images. In this presentation, we resize these 900 variables into shape 30 by 30. In other words, we get a smaller matrix that we can plot that exhibit mini version of similar patterns as before. We use Model 4 to generated the predicted probabilities. These probabilities are printed on the top left corner of each image and they are color coded similarly as before (red probabilities have ground truth of COVID class while green probabilities have ground truth of non-COVID class).

\subsection{Conclusions}
\textbf{Explainable AI System for Early COVID-19 Screening} As the most important contribution of this paper, an Explainable Artificial Intelligence (XAI) system is proposed to assist radiologists for the initial screening of COVID-19 and other related diseases using chest X-ray images for treatment and disease control. This innovation can revolutionize the application of how AI systems are deployed in hospitals and healthcare systems. We anticipate that other related diseases with viral pneumonia signs can use the same detection methods proposed in our paper, which ensure the development of testing procedures with accountability, responsibility, and transparency to human users and patients.

\textbf{A Heuristic and Theoretical Framework of XAI.} This paper introduces a heuristic and theoretical framework for addressing the XAI problems in large-scale and high-dimensional data sets. \textbf{We provide three dimensions as necessary conditions and premises required for a measure to be regarded as explainable and interpretable.} The first dimension, $\mathcal{D}1$, states an interpretable measure does not need to rely on the knowledge of the true model, because any mistakes made in model fitting would be carried over in explaining the features. The second dimension, $\mathcal{D}2$, describes that an interpretable measure should be able to indicate the impact a combination of variables have on the response variable. This means that any inclusion of influential variables would increase this measure while any injection of noisy and useless variables would decrease this measure. This desirable property allows human users to directly make comparisons of the impact the features have when any classifier is trained to make prediction decisions. Though the paper provided detailed work with an arbitrary image data set, the proposed method can be generalized and adapted to any big-data problems. Moreover, it opens up future pipelines for feature selection and dimension reduction in any large-scale and high-dimensional data sets. Last, the third dimension, $\mathcal{D}3$, associates an interpretable measure with the predictivity of a sets of features. This helpful property benefits human users, because it allows us to establish connections and foresee the potential prediction performance (such as AUC values) that a set of features can deliver before any model fitting procedure.

\textbf{An Interaction-based Convolutional Neural Network (ICNN).} To address the XAI problems heuristically described above, this paper introduced a novel design of an explainable and self-interpretable Interaction-based Convolutional Neural Network (ICNN). \textbf{We provide a flexible approach to contribute to the major issues about explainability, interpretability, transparency, and trustworthiness in black-box algorithms. We introduce and implement a non-parametric and interaction-based feature selection methodology and use this as replacement of pre-defined filters that are widely used in ultra-deep CNNs. Under this paradigm, we present an Interaction-based Convolutional Neural Network (ICNN) that extract important features. The proposed architecture uses these extracted features to construct influential and predictive variable modules that are directly associated with the predictivity of the response variable. The proposed design and its many prudent characteristics provide an extremely flexible pipeline that can learn, extract useful information, and unleash the hidden potential from any large-scale or high-dimensional data sets.} The proposed methods have been presented with both artificial examples and real data application in the COVID-19 Chest X-ray Image Data. We conclude from both simulation and empirical application results that I-score shows unparalleled potential to explain informative and influential local information in large-scale data set. The high I-score values suggest that local information possess capability to have higher lower bounds of the predictivity which thus leads to not only high prediction performance but also explanatory power. By arranging features according to I-score from high to low, we are able to cater the dimension of our model to any neural network architecture. Furthermore, we have also shown undiscovered field of an interaction-based neural network architecture which can help us move towards Explainable Artificial Intelligence many steps closer. In fact, we believe that the proposed design can be adapted in any type of CNN. Thus, any CNN architecture that adapts the proposed technology can be regarded as Interaction-based Convolutional Neural Network (ICNN or Interaction-based Network). We encourage both the statistics and computer science communities to further explore this area to deliver more transparency, trustworthiness, and accountability to deep learning algorithms and to build a world with truly Responsible A.I.. 

\end{document}